\RequirePackage{amsthm}

\documentclass[pdflatex,sn-apa]{sn-jnl}

\usepackage{graphicx}%
\usepackage{multirow}%
\usepackage{amsmath,amssymb,amsfonts}%
\usepackage{amsthm}%
\usepackage{mathrsfs}%
\usepackage[title]{appendix}%
\usepackage{xcolor}%
\usepackage{textcomp}%
\usepackage{manyfoot}%
\usepackage{booktabs}%
\usepackage[linesnumbered, ruled]{algorithm2e}
\usepackage{algpseudocode}%
\usepackage{listings}%
\usepackage{enumitem}
\usepackage{subfig}
\usepackage[defaultcolor=red]{changes}
\usepackage{lmodern}
\usepackage{float}

\theoremstyle{thmstyleone}%
\theoremstyle{thmstyletwo}%

\theoremstyle{thmstylethree}%

\begin{document}

\title[Transforming Credit Risk Analysis: A Time-Series-Driven ResE-BiLSTM Framework for Post-Loan Default Detection]{Transforming Credit Risk Analysis: A Time-Series-Driven ResE-BiLSTM Framework for Post-Loan Default Detection}


\author[1]{\fnm{Yue} \sur{Yang}}\email{scxyy2@nottingham.edu.cn}

\author[1]{\fnm{Yuxiang} \sur{Lin}}\email{ssyyl35@nottingham.edu.cn}

\author[2]{\fnm{Ying} \sur{Zhang}}\email{smyyz22@nottingham.edu.cn}

\author[2]{\fnm{Zihan} \sur{Su}}\email{smyzs4@nottingham.edu.cn}

\author[1]{\fnm{Chang Chuan} \sur{Goh}}\email{scxcg1@nottingham.edu.cn}

\author[1]{\fnm{Tangtangfang} \sur{Fang}}\email{scytf1@nottingham.edu.cn}

\author*[1]{\fnm{Anthony Graham} \sur{Bellotti}}\email{anthony-graham.bellotti@nottingham.edu.cn}

\author*[1]{\fnm{Boon Giin} \sur{Lee}}\email{boon-giin.lee@nottingham.edu.cn}

\affil[1]{\orgdiv{School of Computer Science}, \orgname{University of Nottingham Ningbo China}, \city{Ningbo}, \postcode{315100}, \state{Zhejiang}, \country{China}}

\affil[2]{\orgdiv{Department of Mathematical Sciences}, \orgname{University of Nottingham Ningbo China}, \city{Ningbo}, \postcode{315100}, \state{Zhejiang}, \country{China}}


\abstract{Prediction of post-loan default is an important task in credit risk management, and can be addressed by detection of financial anomalies using machine learning. This study introduces a ResE-BiLSTM model, using a sliding window technique, and is evaluated on 44 independent cohorts from the extensive Freddie Mac US mortgage dataset, to improve prediction performance. The ResE-BiLSTM is compared with five baseline models: Long Short-Term Memory (LSTM), BiLSTM, Gated Recurrent Units (GRU), Convolutional Neural Networks (CNN), and Recurrent Neural Networks (RNN), across multiple metrics, including Accuracy, Precision, Recall, F1, and AUC. An ablation study was conducted to evaluate the contribution of individual components in the ResE-BiLSTM architecture. Additionally, SHAP analysis was employed to interpret the underlying features the model relied upon for its predictions. Experimental results demonstrate that ResE-BiLSTM achieves superior predictive performance compared to baseline models, underscoring its practical value and applicability in real-world scenarios.}

\keywords{Credit risk; machine learning; loan prediction; anomaly detection}



\maketitle

\section{Introduction}\label{sec1}

The significance of anomaly detection lies in its role in identifying unusual patterns in complex data, thus mitigating potential risks in various fields such as machine failure, financial fraud, web error logs, and medical health diagnosis \citep{2021_16}. In financial fraud detection, fraudulent activities usually fall into four categories \citep{Al-Hashedi_2021}: banking, corporate, insurance, and cryptocurrency fraud. Banking fraud includes credit card, loan, and money laundering fraud. Corporate fraud consists of financial statement fraud, securities and commodities fraud, while insurance fraud involves life and auto insurance fraud.

This paper focuses on anomaly detection in loans. The detection of financial loan anomalies consists of two primary phases: pre-loan fraud detection, known as the ``application model'', and post-loan default prediction, identified as the ``behavioral model'' \citep{2020_17}. The purpose of pre-loan fraud detection is to intercept fraudulent activities during the loan application process \citep{2020_16}, usually based on upfront audits \citep{2020_13}. These activities can involve falsifying financial or identity information and misrepresenting intentions. In contrast, using data analytics and machine learning, post-loan default prediction assesses risk by analyzing historical financial data of the borrowers \citep{2019_5}. This prediction assists financial institutions in implementing preventative strategies and modifying loan conditions to reduce non-performing loans, thereby ensuring asset quality assurance and stability.

Financial loan data, typically documented on a monthly basis as time series data. The distinctive feature of time-series data is the temporal dependency between data points, which requires consideration of this temporal dependence during anomaly detection. This complexity makes anomaly detection in time series data more difficult compared to static data that do not involve time\citep{2021_9}.  Currently, the most commonly used and state-of-the-art methods for handling time series are based primarily on long-short-term memory (LSTM) models \citep{Pardeshi_2023,Gao_2023}. LSTM is effective for time series, but its unidirectional design does not efficiently manage bidirectional dependencies, leading to reduced sample recognition. The Bidirectional Long Short-Term Memory Network (BiLSTM) model was introduced \citep{Siami_2019} to address this issue, combining forward and backward LSTM unit layers to capture dependencies simultaneously in time series data \citep{2019_8}, thus offering a richer understanding of temporal patterns. This model is proficient in identifying intricate time series characteristics, improving the prediction of post-loan default. However, as sequential models, both BiLSTM and LSTM models do not possess robust global modeling capabilities. Furthermore, there are limited studies using post-loan approval repayment behavior to design default models. 

Moreover, one significant reason loan prediction models are rarely used in real-life scenarios is that algorithms such as LSTM are considered black-box models, which can lead to trust issues \citep{2021_18}. Therefore, providing clear causal explanations for the model's predictions is crucial to ensuring financial security. To address this limitation, Explainable AI (XAI) has emerged as a promising approach, enhancing the interpretability and accountability of ML models while ensuring that human users can comprehend the reasoning behind predictions \citep{2023_43}. 

This study proposes ResE-BiLSTM, a BiLSTM model integrated with a Residual-enhanced Encoder for feature extraction, to improve out-of-sample (OOS) prediction accuracy. It is applied to the Freddie Mac Single-Family Loan-Level Dataset \citep{FreddieMac} and uses SHapley Additive exPlanations (SHAP) \citep{2017_2}, a widely used XAI method for models based on LSTM \citep{2023_41}, to explain the differences in prediction basis between the proposed model and the baseline model. The key contributions of this paper are:
\begin{enumerate}
    \item To propose the design of the ResE-BiLSTM model for time-series-based default prediction tasks and conduct a comprehensive evaluation..

    \item To perform a performance evaluation of the ResE-BiLSTM model against state-of-the-art studies in post-loan default prediction.
    
    \item To perform analysis on how the predictions of the proposed and baseline models are affected by the importance of the characteristics of the time-series data using SHAP.
    
\end{enumerate}

\section{Related Works}\label{sec2}

\subsection{Benchmark Datasets and Loan Default Prediction Model}\label{subsec2-1}


Most prior studies have validated the performance of the default model using publicly accessible benchmark datasets from two primary sources, including Freddie Mac \citep{FreddieMac} and Lending Club \citep{LendingClub}. In contrast, some studies used private datasets, making reproduction or replication of their results challenging for research purposes.

Specifically, \cite{Zandi_2024} introduced dynamic multi-layer graph neural networks (DYMGNN) using the Freddie Mac dataset, achieving a loan default prediction F1 score of 0.851. \cite{Wang_2024} adopted a survival model combined with neural networks on the same dataset, providing an interpretable model that elucidates the risk of default with factors such as loan maturity, origination year and environmental influences. \cite{Karthika_2023} developed an XGBoost-based BiGRU with a self-attention mechanism (XGB-BiGRU-SAN), achieving more than 98\% mean precision, precision and recall on the Freddie Mac and Lending Club datasets. \cite{2023_29} reported 89\% accuracy using a logistic regression model, 78\% with ridge regression, and 76\% with k-nearest neighbors to predict loan prepayment, a bank risk indicator for mortgage-backed securities (MBS), using the Freddie Mac dataset.

However, these studies did not consider monthly Freddie Mac repayment data, thus neglecting borrower behavior in a substantial historical dataset.

\subsection{Design of BiLSTM and Its Variants in Anomaly Detection}\label{subsec2-2}


Recent research has used BiLSTM models to detect financial anomalies, typically integrating them with various mechanisms such as attention, convolutional neural networks (CNN), and Transformer networks.

\cite{Chen_2022} used an attention-based BiLSTM model to analyze the data sequences to discover contract flaws with an accuracy of 95.40\% and an F1 score of 95.38\% compared to baseline models such as LSTM, GRU and CNN. \cite{R_2024} utilized a similar model structure for detecting credit card fraud, where BiLSTM was used for feature extraction followed by an attention layer, forming the A-BiLSTM algorithm, which achieved 99.96\% accuracy on the European Credit Card dataset, which is better than its baseline models LSTM and BiLSTM. \cite{Narayan_2022} introduced a Hybrid Sampling (HS) - Similarity Attention Layer (SAL) - BiLSTM method to improve the classification performance in the detection of credit card fraud by removing redundant samples from the majority class and adding instances to the minority class.

Several studies analyzed the integration of BiLSTM, attention, and CNN for financial anomaly detection. \cite{Agarwal2024} introduced a CNN-BiLSTM-Attention where CNN handles data initially, BiLSTM provides historical context next, and the attention mechanism discerns transaction multicollinearity, tested with 97\% recall in IEEE-CIS Fraud Detection Dataset. \cite{Joy_2023} presented a BiLSTM and CNN model driven by the attention mechanism, improving feature extraction and classification, outperforming CNN and BiLSTM-with-CNN on the Talking Data dataset. \cite{K_2023} developed a structure that effectively uses CNN for feature extraction and BiLSTM for sequence learning, with the focus on words. This model improves Korean voice phishing detection with 99.32\% accuracy and a 99.31\% F1 score, which outperforms the CNN, LSTM, and BiLSTM baselines.

Several studies have proposed the integration of BiLSTM with Transformer networks, where Transformer, an algorithm based on the multi-headed self-attention mechanism introduced by \cite{transformer}, is capable of capturing long-range contextual information across the entire sequence. \cite{Cai_2021} developed a hybrid model with BiLSTM and Transformer to improve sentiment classification. Initially, BiLSTM derives contextual features, which are trained in several independent Transformer modules. The parameters of each Transformer are optimized during training to precisely determine sentiment polarity. Experiments on the SemEval dataset showed that this model outperforms traditional models such as CNN, LSTM, and BiLSTM in sentiment classification. \cite{Boussougou_2023} applied a similar approach, integrating Transformer and BiLSTM for portfolio return prediction. The input data fed into a post-BiLSTM three-layer encoder to produce predicted outputs, demonstrating the effectiveness of the BiLSTM-Transformer model in portfolio return prediction.

LSTM, GRU, CNN, and RNN are commonly used standard models in time series data-based anomaly detection studies \citep{2023_21}. LSTM networks, with their gated design, effectively handle the problem of vanishing gradients seen in traditional RNNs, making them useful for capturing long-term dependencies in sequence data. CNNs are adept at detecting local patterns in sequences and are often combined with RNN models to improve spatio-temporal pattern tasks. The GRU, a simplified version of RNN, simplifies the gating functions of LSTM, offering similar performance with faster training, and is popular for anomaly detection and forecasting in time-series data \citep{2024_7}.

\subsection{XAI in Loan Default Prediction}\label{subsec2-3}

\cite{2023_39} defines XAI as ``AI systems that can explain their reasoning to humans, indicate their strengths and weaknesses, and predict their future behavior''. Unlike traditional ``black-box'' models, XAI offers insight into the internals of complex models, improving credibility and helping to comply with regulatory requirements in sectors such as finance, healthcare and law enforcement. XAI covers an array of methods designed for different objectives, offering various levels of insight. These methods are generally divided into pre-model, in-model, and post-model techniques \citep{2023_41}. The post-model technique, such as SHAP, Local Interpretable Model-Agnostic Explanations (LIME) \citep{2023_43}, and Partial Dependence Plots (PDP) \citep{2023_43}, is frequently utilized to clarify results from pre-trained models. 

SHAP, derived from cooperative game theory \citep{2017_2}, evaluates the impact of each input feature on the output of the model by assigning importance scores, highlighting the influential input features in the predictions. Conversely, LIME makes small data perturbations and builds an interpretable surrogate model to approximate the behavior of the black-box model. PDP shows how the values of a single input feature influence the predictions on average, which is explained by its global effect. Each method is suitable for specific domains. \cite{2023_41} cataloged anomaly explanation techniques over 22 years, advocating for selection based on the model type. In particular, LSTM models often use SHAP to explain anomalies where the study by \cite{2021_18} indicated that LIME offers slightly better interpretability than SHAP in the detection of credit card fraud.

Decision trees, linear regression, and rule-based classifiers are inherently interpretable in-model techniques due to their straightforward structures, allowing for transparency via human-readable decision rules or coefficients, directly correlating predictions with input features. In loan default prediction, these models can elucidate the impact of borrower behavior or demographic factors on default risk. However, there is a balance between interpretability and predictive accuracy, as highly interpretable models often do not perform optimally \citep{2021_19}. \cite{2023_38} demonstrated that pre-model strategies improve transparency in data preprocessing by employing an X-LSTM model with SHAP or LIME to identify crucial training features, with results documented on a blockchain. This method streamlines input data, improves performance and interpretability, and reveals key predictive features.

In general, the use of XAI in predicting loan defaults improves the transparency of the model that could help build trust between financial institutions, aligns with regulatory standards, and optimizes model performance. Choosing suitable XAI tools based on specific application contexts effectively clarifies model decisions, guaranteeing the wide applicability of these models. For complex deep learning models, SHAP is mainly used to provide intuitive feature contribution values, aiding in understanding model decisions. Thus, this study uses SHAP as the interpretability method for model explanation.

\section{Methodology}\label{sec3}

\subsection{Data Preprocessing}\label{subsec3-1}
This study uses the Freddie Mac Single-Family Loan-Level Dataset, involving more than 50 million entries from 1999 onward. Due to the reduced significance of older data and the incomplete recent data, the study focuses on monthly repayment data from loans available between 2009 and 2019. Each quarter constitutes a separate dataset, with the first 1,000,000 records selected from each. Table \ref{data} displays the basic statistics of the 44 cohorts: number of loans, average and median loan history length, and default rate. Although these datasets are chronologically ordered, they are independent and represent the repayment records from a specific quarter across all following years to the present. The selected features are shown in Table\ref{feature}.

\begin{table}[!ht]
\centering
\caption{Summary of the 44 independent cohorts in the Freddie Mac Single-Family Loan-Level Dataset}
\label{data}
\footnotesize
\begin{tabular*}{\textwidth}{@{\extracolsep\fill}lcccc}
\toprule
\multirow{2}*{\textbf{Cohort}} & \textbf{Number of} & \textbf{Average Loan} & \textbf{Median Loan} & \textbf{Default} \\ 
& \textbf{Loans} & \textbf{Length} & \textbf{Length} & \textbf{Rate} \\ \midrule
2009Q1   & 17604        & 56.805              & 42   & 1.755\%       \\
2009Q2   & 16730        & 59.773              & 42   & 1.470\%       \\
2009Q3   & 15728        & 63.581              & 44   & 2.893\%       \\
2009Q4   & 16080        & 62.189              & 42   & 2.674\%       \\
2010Q1   & 15779        & 63.375              & 42   & 2.884\%       \\
2010Q2   & 16132        & 61.989              & 40   & 3.149\%       \\
2010Q3   & 15525        & 64.412              & 46   & 2.209\%       \\
2010Q4   & 12957        & 77.178              & 67   & 1.443\%       \\
2011Q1   & 13969        & 71.587              & 61   & 2.098\%       \\
2011Q2   & 16211        & 61.687              & 47   & 2.807\%       \\
2011Q3   & 15196        & 65.807              & 55   & 2.165\%       \\
2011Q4   & 12631        & 79.170              & 76   & 1.362\%       \\
2012Q1   & 12304        & 81.274              & 85   & 1.756\%       \\
2012Q2   & 11566        & 86.460              & 92   & 1.816\%       \\
2012Q3   & 11209        & 89.214              & 95   & 1.963\%       \\
2012Q4   & 10939        & 91.416              & 97   & 1.901\%       \\
2013Q1   & 11138        & 89.783              & 96   & 2.182\%       \\
2013Q2   & 11444        & 87.382              & 92   & 2.386\%       \\
2013Q3   & 12733        & 78.536              & 83   & 2.592\%       \\
2013Q4   & 15045        & 66.467              & 65   & 2.951\%       \\
2014Q1   & 16002        & 62.492              & 62   & 3.412\%       \\
2014Q2   & 16287        & 61.399              & 61   & 2.923\%       \\
2014Q3   & 15715        & 63.633              & 67   & 2.660\%       \\
2014Q4   & 15778        & 63.379              & 66   & 2.903\%       \\
2015Q1   & 15638        & 63.947              & 67   & 2.954\%       \\
2015Q2   & 14592        & 68.531              & 68   & 2.947\%       \\
2015Q3   & 15923        & 62.802              & 63   & 3.410\%       \\
2015Q4   & 15860        & 63.052              & 62   & 3.140\%       \\
2016Q1   & 17180        & 58.207              & 59   & 3.423\%       \\
2016Q2   & 16102        & 62.104              & 59   & 3.198\%       \\
2016Q3   & 15871        & 63.008              & 60   & 3.459\%       \\
2016Q4   & 15961        & 62.653              & 61   & 3.390\%       \\
2017Q1   & 19386        & 51.584              & 49   & 4.354\%       \\
2017Q2   & 20066        & 49.836              & 45   & 4.625\%       \\
2017Q3   & 20019        & 49.953              & 44   & 4.311\%       \\
2017Q4   & 20194        & 49.520              & 44   & 4.600\%       \\
2018Q1   & 22999        & 43.480              & 39   & 4.913\%       \\
2018Q2   & 26343        & 37.961              & 31   & 4.555\%       \\
2018Q3   & 29629        & 33.751              & 27   & 4.398\%       \\
2018Q4   & 32095        & 31.158              & 24   & 4.518\%       \\
2019Q1   & 36704        & 27.245              & 22   & 4.795\%       \\
2019Q2   & 34226        & 29.218              & 22   & 4.885\%       \\
2019Q3   & 32061        & 31.191              & 24   & 4.429\%       \\
2019Q4   & 30083        & 33.241              & 29   & 4.168\%      	\\
\bottomrule
\end{tabular*}
\end{table}

\begin{table}[h]
\small
\centering
\caption{Overview of the features in the Freddie Mac Single-Family Loan-Level Dataset.}
\begin{tabular*}{\textwidth}{@{\extracolsep\fill}llp{5cm}}
\toprule%
\textbf{No.} & \textbf{Feature} & \textbf{Description} \\
\midrule
1  & Loan Sequence Number & Unique ID allocated for every loan.\\
2  & Current Actual UPB & Indicates the reported final balance of the mortgage.\\
3  & Current Loan Delinquency Status & Days overdue relative to the due date of the most recent payment made. \\
4  & Defect Settlement Date & Date for resolution of Underwriting or Servicing Defects that are pending confirmation. \\
5  & Modification Flag & Signifies that the loan has been altered.\\
6  & Current Interest Rate (Current IR) & Displays the present interest rate on the mortgage note, with any modifications included. \\
7  & Current Deferred UPB & The current non-interest bearing UPB of the modified loan.\\
8 & Due Date Of Last Paid Installment (DDLPI) & 
Date until which the principal and interest on a loan are paid.\\
9 & Estimated Loan To Value (ELTV) & LTV ratio using Freddie Mac's AVM value.\\
10 & Delinquency Due To Disaster & Indicator for hardship associated with disasters as reported by the Servicer.\\
11 & Borrower Assistance Status Code & Type of support arrangement for interim loan payment mitigation.\\
12 & Current Month Modification Cost & Monthly expense resulting from rate adjustment or UPB forbearance. \\
13 & Interest Bearing UPB & The interest-bearing UPB of the adjusted loan. \\
\bottomrule
\label{feature}
\end{tabular*}
\end{table}

The feature selection process involves removing features that mainly exhibit missing values and those linked to categorical attributes. The Current Loan Delinquency Status acts as the class label, where a value of 3 or more signifies that the borrower has not repaid the loan for at least 3 months, considering it a default, aligning with the industry standard definition according to Basel II guidelines \citep{BCBS2006}. Discrete features are transformed using hot encoding. Two additional features are introduced, including the difference in Interest Bearing UPB and Current Actual UPB between the current month and the previous month. These differences generate new features labeled Interest Bearing UPB Delta and Current Actual UPB Delta, respectively.

Interest Bearing UPB, or Interest Bearing Unpaid Principal Balance, signifies the portion of a modified mortgage's unpaid principal balance subject to interest. This amount is the basis for interest calculations and represents the remaining owed balance of a borrower. Calculating Interest Bearing UPB-Delta is valuable in loan default prediction and financial modeling, as it offers insights into repayment patterns. A negative value suggests principal repayment, indicating normal behavior, whereas a zero value might signal missed payments, indicating risk. A positive increase in principal may result from loan restructuring, deferred capitalization, or new debt, which requires further investigation.

The Current Actual UPB, combining both interest-bearing and non-interest-bearing UPB, offers a complete view of the borrower's debt. This metric is important for risk management and thorough loan evaluation. The feature Current Actual UPB-Delta, indicating changes in deferred principal, adds further time-series insights by capturing adjustments such as additions, reductions, or re-amortizations. These elements improve the model's capacity to differentiate typical repayment behavior from the distinct patterns linked to loan modifications.

The data is then organized by Loan Sequence Number (the loan ID). Within each group, the sliding window \citep{2023_26} is applied with a window length of 19 months as recommended by existing studies \citep{window1,window2,window3,window4,window5}.  Each 19-month slice was divided into three parts: the first 14 months served as input features for the model, months 15 to 16 were designated as a blank period, and the final 3 months were used as the observation window for generating labels. The data is then randomly divided into 70\% as the training set and 30\% as the testing set (out-of-sample test) according to the original default ratio. To prevent data leakage, time slices from the same user were ensured not to appear in both the training and test sets. According to the definition of default adopted in this study, a default is identified when $\text{CLDS} \geq 3$ occurs within the label window. Accordingly, a label of $y$ = 1 is assigned if such an event occurs during the observation period; otherwise, the label is set to $y$ = 0. To further mitigate potential label leakage arising from early warning signals (such as CLDS = 1 or 2) appearing in the input window, any samples with nonzero CLDS values in the first 14 months were removed. As a result, defaults ($\text{CLDS} \geq 3$) would not occur during the blank period (months 15–16), and true default events could only begin from month 17 onward. To address class imbalance, random undersampling was applied to the training set, resulting in a 1:1 ratio between default and non-default time slices.

\subsection{Proposed ResE-BiLSTM Model}\label{subsec3-2}

\begin{figure}[!ht]
    \centering
    \includegraphics[width=.55\textwidth]{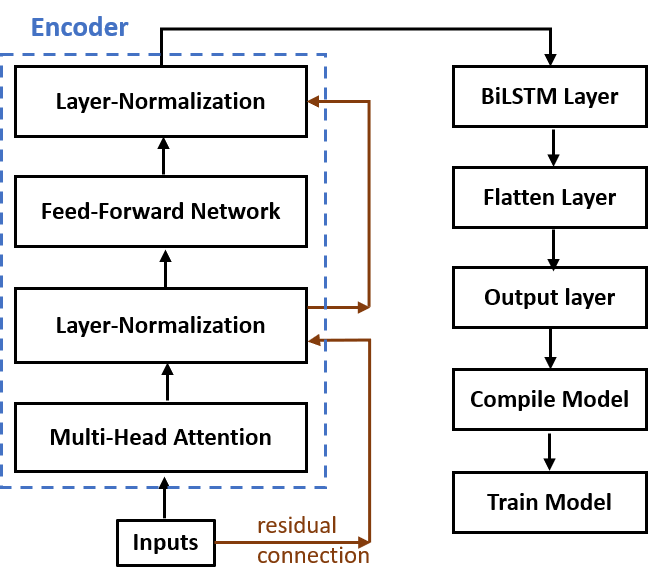}
    \caption{\centering Overview design of the proposed ResE-BiLSTM model.}
    \label{fig:model_structure}
\end{figure}

Figure \ref{fig:model_structure} illustrates the ResE-BiLSTM model, a hybrid deep learning architecture designed for loan default prediction. It combines a Residual-enhanced Encoder with Bidirectional Long Short-Term Memory Networks (BiLSTM) to effectively capture temporal dependencies and improve model performance. As shown, the model first utilizes a multi-head attention mechanism, which serves to focus on the most relevant features within the time-series data, followed by a Feedforward Neural Network (FNN) that forms the encoder, enabling the model to learn richer representations. The output from the encoder is then passed into the BiLSTM layer, which captures both forward and backward dependencies in the time sequence.

In addition to these components, the ResE-BiLSTM architecture incorporates residual connections, which help mitigate the vanishing gradient problem and enhance the flow of information across layers, improving model stability and convergence. The model handles input data of dimensions $(T, F)$, with $T$ as the sequence length and $F$ the number of features.

The pseudocode for this model is presented in Algorithm \ref{alg:prediction}, which outlines the detailed process for feature extraction, temporal dependency modeling, and prediction.

\begin{algorithm}[!ht]
  \KwIn{List $X = (X_1, X_2, ..., X_t)$}
  \KwOut{predictedValue $\bar{y}$}
  
  \tcp{Multihead Attention}
  Initialize $Q = XW^Q$, $K = XW^K$, $V = XW^V$\
  
  \For{$i=1$ \KwTo $h$}{
      $A_i = \text{softmax}\left(\frac{Q_i K_i^T}{\sqrt{d_k}}\right)$\\
      $Z_i = A_i V_i$\
  }
  
  $Z = \text{Concat}(Z_1, Z_2, ..., Z_h)$\\
  $X_o = ZW^O$\\
  $ \text{NormX}_o = \text{LayerNormalization}(X_o, X)D$\\
  $ff_X = \text{feedForwardLayer}(\text{NormX}_o)$\\
  $X =  \text{LayerNormalization}(ff_X,X_o)$\\

  \tcp{BiLSTM}
  Initialize $h_f(0) = h_0$, $c_f(0) = 0$\\
  
  \For{$t=1$ \KwTo $T$}{
      $f_t^{(f)} = \sigma(W_f^{(f)} [X_t,h_f(t-1)] + b_f^{(f)})$\\
      $i_t^{(f)} = \sigma(W_i^{(f)} [X_t,h_f(t-1)] + b_i^{(f)})$\\
      $\tilde{C}_t^{(f)} = \tanh(W_c^{(f)} [X_t, h_f(t-1)] + b_c^{(f)})$\\
      $o_t^{(f)} = \sigma(W_o^{(f)} [X_t ,h_f(t-1)] + b_o^{(f)})$\\
      $c_f(t) = f_t^{(f)} c_f(t-1) + i_t^{(f)} \tilde{C}_t^{(f)}$\\
      $h_f(t) = o_t^{(f)} \tanh(c_f(t))$\\
  }
  
  \For{$t=T$ \upshape{\textbf{down to}} $1$}{
      $f_t^{(b)} = \sigma(W_f^{(b)} [X_t , h_b(t+1)] + b_f^{(b)})$\\
      $i_t^{(b)} = \sigma(W_i^{(b)} [X_t, h_b(t+1)] + b_i^{(b)})$\\
      $\tilde{C}_t^{(b)} = \tanh(W_c^{(b)} [X_t,h_b(t+1)] + b_c^{(b)})$\\
      $o_t^{(b)} = \sigma(W_o^{(b)} [X_t ,h_b(t+1)] + b_o^{(b)})$\\
      $c_b(t) = f_t^{(b)} c_b(t+1) + i_t^{(b)} \tilde{C}_t^{(b)}$\\
      $h_b(t) = o_t^{(b)} \tanh(c_b(t))$\
  }
  
  $h = ([h_f(1), h_b(1)], [h_f(2), h_b(2)], ..., [h_f(T), h_b(T)])$\\
  $ \text{FlattenH} = \text{Flatten}(h)$\\
  $\bar{y} = \text{FullyConnectedLayer}(\text{FlattenH})$\\

  \caption{Pseudocode of the proposed ResE-BiLSTM.}
  \label{alg:prediction}
\end{algorithm}

\subsubsection{Residual-enhanced Encoder (ResE) Layer}\label{subsubsec3-2-1}
\begin{enumerate}[label=\Alph*.]

\item Multi-Head Attention \\

Multi-head attention \citep{vaswani2017attention} is a sophisticated attention mechanism integrating several attention processes in one model. It functions by projecting input into multiple subspaces via linear transformations with learned weight matrices. Each head processes its own transformed input independently, allowing the model to concentrate on different data aspects and grasp richer contextual details. This model utilizes a self-attention mechanism that computes attention using only the input, without external data. This method efficiently captures relationships and dependencies within the input sequence. Importantly, post-attention calculation maintains the output dimensionality consistent with the input, facilitating integration with subsequent layers.

The query vector $ Q $, key vector $ K $, and value vector $ V $ are initially derived from linear transformations, with $ Q = XW^Q $, $ K = XW^K $, and $ V = XW^V $, where $ X $ represents the input data and $ W^Q, W^K, W^V $ are weight matrices randomly initialized. The algorithm utilizes $ h $ attention heads to derive the attention matrix $ A_i $ via the scaled dot product $ \frac{Q_i K_i^T}{\sqrt{d_k}} $, where $ d_k $ denotes the key vector dimension, and subsequently employs the softmax function to produce a probability distribution. The output $ Z_i $ for each attention head is derived by applying the attention weights $ A_i $ to the value vector $ V_i $. Finally, combining the outputs from all attention heads and projecting back into the input space with $ W^{O} $ completes the multi-head attention layer. \\

\item Normalization Layer and Residual Connection Mechanism \\

The normalization layer follows the multi-head attention and feed-forward network to improve model stability and performance. The residual connection in layer normalization ensures balanced input and layer output contributions \citep{2016_1}, preserving essential information from earlier layers and enabling deeper layers to learn more complex features. Moreover, normalization layer mitigates issues like vanishing and exploding gradients through output standardization, improving training stability. It also reduces the influence of input scale variations on parameter updates, speeding up convergence and optimizing the efficiency of the training process. The normalization layer operates as follows:

\begin{equation}
    LN(x) = \frac{x - \mu}{\sqrt{\sigma^2 + \epsilon }}\gamma + \beta
\end{equation}

\noindent where $\mu$ represents the mean of each feature, $\sigma^2$ indicates their variance, $\epsilon$ is a small constant (set at $1\text{e-}6$) to avoid division by zero, as well as $\gamma$ and $\beta$ are learnable parameters. 

The residual connection mechanism incorporated within the ResE module plays a pivotal role in facilitating effective deep representation learning. Specifically, by introducing skip connections that directly add the input of a sub-layer (e.g., the multi-head attention or feed-forward layer) to its output prior to normalization, the model preserves the integrity of the original feature representations while enabling the training of deeper networks without degradation. This architectural design mitigates the vanishing gradient problem and ensures more stable and efficient gradient flow during backpropagation. Moreover, the integration of residual connections with layer normalization enhances the model’s capacity to learn complex temporal dependencies by stabilizing the output distributions across layers.\\

\item Feed-Forward Network \\

The feed-forward network processes each time step independently, refining and improving the fine-grained features to improve feature representation \citep{2024_8}. Using the ReLU activation function, the network applies non-linear transformations to capture more intricate patterns and relationships within the data. The feed-forward network in this model features two layers. The initial layer is fully connected, containing 256 neurons and employing ReLU activation. The second layer reshapes the feature dimension to match the original input, maintaining compatibility with the BiLSTM layer. The resultant output is shaped as batch size, sequence length, feature dimension. This is followed by layer normalization applied to the combined outputs of the feed-forward network and attention layer, improving stability and robustness of the model.

\end{enumerate}

\subsubsection{BiLSTM}\label{subsubsec3-2-2}

BiLSTM processes time series data bidirectionally, capturing temporal relationships and contextual information \citep{schuster1997bidirectional}. Equipped with forget, input, and output gates, it selectively retains and updates information to capture dependencies, enhancing its applicability to predict loan default, where temporal patterns are crucial. Moreover, BiLSTM complements the Residual-enhanced Encoder by learning local time-series patterns, while the Residual-enhanced Encoder captures global dependencies. This combination promotes robust data representation.

The BiLSTM algorithm manages sequence processing through two states: the cell state ($ c $) for long-term memory and the hidden state ($ h $) for short-term context and time-step output. It operates bidirectionally over the sequence, forward from $ t=1 $ to $ t=T $ and backward from $ t=T $ to $ t=1 $. At each step, the hidden forward and backward states ($ h_f $ and $ h_b $) are concatenated to form the final output, integrating the dependencies of the past and future sequences. For process initialization, $ c $ and $ h $ of both forward LSTM ($ c_f(0), h_f(0) $) and backward LSTM ($ c_b(T+1), h_b(T+1) $) start as zero vectors.

\begin{enumerate}[label=\Alph*.]

\item Forward LSTM Process \\

The LSTM executes these operations at every time step: \\

\begin{enumerate}
    \item \textbf{Forget Gate}

    This mechanism determines which part of the previous cell state ($ c_f(t-1) $) is preserved in the cell state, defined as follows:
    \[
    f_t^{(f)} = \sigma(W_f^{(f)} [X_t,h_f(t-1)] + b_f^{(f)})
    \]
    where $ \sigma $ is the sigmoid activation function mapping values to [0, 1], $ W_f^{(f)} $ represents the weights for the forward forget gate, $ X_t $ is the input data, $ h_f(t-1) $ denotes the prior hidden state, and $ b_f^{(f)} $ is the forget gate bias. \\

    \item \textbf{Input Gate}

    This operation determines the portion of current input ($ X_t $) to be stored in the cell state, defined as follows:
    \[
    i_t^{(f)} = \sigma(W_i^{(f)} [X_t,h_f(t-1)] + b_i^{(f)})
    \]
    where $ W_i^{(f)} $ is the weights for the forward input gate, $ h_f(t-1) $ is the hidden state from the preceding time step, $ b_i^{(f)} $ is the bias term for the input gate. \\
    
    \item \textbf{Candidate Cell State}

    This process calculates a candidate value ($ \tilde{C}_t^{(f)} $) for potentially updating the cell state as follows:
    \[
    \tilde{C}_t^{(f)} = \tanh(W_c^{(f)} [X_t, h_f(t-1)] + b_c^{(f)})
    \]
    where $ W_c^{(f)} $ represents the weights for the candidate cell state, $ h_f(t-1) $ is the previous time step's hidden state, and $ b_c^{(f)} $ is the bias term for the candidate cell state. \\

    \item \textbf{Output Gate}

    This operation delineates the cell state fraction impacting the hidden state ($ h_f(t) $), defined as follows:
    \[
    o_t^{(f)} = \sigma(W_o^{(f)} [X_t ,h_f(t-1)] + b_o^{(f)})
    \]
    where $ W_o^{(f)} $ denotes the forward output gate weights, $ h_f(t-1) $ is the previous hidden state, and $ b_o^{(f)} $ stands for the output gate bias. \\
    
    \item \textbf{Updated Cell State}

    This operation revises the cell state by integrating data from the forget gate, input gate, and candidate cell state, defined as follows:
    \[
    c_f(t) = f_t^{(f)} c_f(t-1) + i_t^{(f)} \tilde{C}_t^{(f)}
    \]
    where $ c_f(t-1) $ denotes the previous cell state, $ f_t^{(f)} $ is the forget gate values, $ i_t^{(f)} $ represents input gate values, and $ \tilde{C}_t^{(f)} $ is the candidate cell state. \\
    
    \item \textbf{Updated Hidden State}

    This function determines the hidden state ($ h_f(t) $) by employing the output gate alongside the new cell state, defined as follows:
    \[
    h_f(t) = o_t^{(f)} \tanh(c_f(t))
    \]
\end{enumerate}

Following these six steps, the forward process refreshes the cell state ($ c_f(t) $) and the hidden state ($ h_f(t) $), producing an output ($ o_t^{(f)} $) through the output gate. \\

\item Backward LSTM Process \\

The backward LSTM functions similarly, but processes in reverse, beginning from $ t = T $ to $ t = 1 $.

\end{enumerate}

\subsubsection{Flatten and Output Layers}\label{subsubsec3-2-3}

The flatten layer transforms the multi-dimensional tensor output from the BiLSTM into a one-dimensional form appropriate for the fully connected layer. Subsequently, a fully connected two-layer network is used for prediction. To limit the output between [0,1], a sigmoid activation function is used in the output layer.

\subsection{Evaluation Metrics}\label{subsec3-3}

This study uses five metrics to evaluate the ResE-BiLSTM model, including accuracy \citep{ref1}, precision \citep{ref26}, recall \citep{ref14}, F1 \citep{ref2}, and area under the ROC curve (AUC) \citep{2022_6}. Accuracy denotes the ratio of correctly predicted samples to the total number of samples. Precision indicates the fraction of true positives among all positive predictions. Recall is the fraction of true positives identified by the model. High recall aids in regulatory compliance, helping banks fully assess risks and enforce suitable controls. Recall and precision have a trade-off; increasing recall tends to reduce precision \citep{LEI2012135}. To counteract precision reduction when recall is maximized, the F1 score, the harmonic mean of precision and recall, is used. The AUC, which varies from 0 to 1, quantifies the ability of a model to differentiate. Values near 1 imply superior performance. These metrics are defined as follows:

\[
\text{Accuracy} = \frac{\text{TP} + \text{TN}}{\text{TP} + \text{TN} + \text{FP} + \text{FN}}
\]

\[
\text{Precision} = \frac{\text{TP}}{\text{TP} + \text{FP}}
\]

\[
\text{Recall} = \frac{\text{TP}}{\text{TP} + \text{FN}}
\]

\[
\text{F1} = 2 \cdot \frac{\text{Precision} \cdot \text{Recall}}{\text{Precision} + \text{Recall}}
\]
\vspace{1em}

Given that different evaluation metrics might yield varying results, using a multi-metric approach ensures a comprehensive model performance evaluation. This study uses $AvgR$ \citep{ref45}, to evaluate the performance of the model on different indicators. Models are initially ranked according to their performance in accuracy, precision, recall, F1 and AUC metrics in various groups (e.g., quarterly or yearly). These rankings are then averaged to obtain the final $AvgR$ (see Section \ref{sec:4.1}) where a lower $AvgR$ indicates better classifier performance.

\section{Experiment Results \& Discussion}\label{sec4}

\subsection{ResE-BiLSTM Model Performance Analysis}\label{sec:4.1}

Tables \ref{tab:A1} through \ref{tab:A5} in the appendix display the average performance of the six models on five metrics, based on 10 independent trials per cohort. The analysis reveals that although individual model performance varied across cohorts, ResE-BiLSTM consistently outperformed all metrics.

ResE-BiLSTM achieved the highest accuracy in 38 cohorts, or 86.36\% of the total, clearly outperforming other models, highlighting the ability of ResE-BiLSTM to capture complex features. In contrast, models such as BiLSTM and GRU showed the best performance in two cohorts, and other models achieved the highest performance in two cohorts. Furthermore, ResE-BiLSTM led in precision, with the highest precision in 26 cohorts, representing 59.09\% of the total, compared to LSTM, BiLSTM, GRU, CNN and RNN, which outperformed in cohorts 1, 3, 4, 1 and 9, respectively.

ResE-BiLSTM achieved the highest recall in 37 cohorts, highlighting its effectiveness in reducing false negatives. Furthermore, it achieved the highest F1 score in 39 out of 44 cohorts (88.64\%), showing an excellent precision-recall balance. Furthermore, the results of the AUC demonstrated that ResE-BiLSTM maintained high true positive rates and low false positive rates in 36 cohorts.

\subsubsection{AvgR Performance Analysis}\label{subsubsec4-1-1}

Let $Y_q$ denote the name of the cohort, where $Y$ indicates the year and $q$ the quarter, such that $2009_1$ stands for the cohort of the first quarter of 2009. Here, $m \in \mathcal{M}$ refers to the model, with $|\mathcal{M}| = 6$. The average ranking of a model $m$ in the $q^{th}$ quarter of year $Y$ is defined as:
\begin{equation}
AvgR_{Y_q}(m) = \frac{1}{5}\big(AccR_{Y_q}(m)+PreR_{Y_q}(m)+RecR_{Y_q}(m)+F1R_{Y_q}(m)+AUCR_{Y_q}(m)\big)
\end{equation}

\normalsize

\noindent where $AccR_{Y_q}(m), PreR_{Y_q}(m), RecR_{Y_q}(m), F1R_{Y_q}(m),$ and $AUCR_{Y_q}(m)$ denote the model $m$ ranking in terms of precision, precision, recall, F1 and AUC metrics in the cohort $Y_q$.\\

Table \ref{table2} shows the average ranking for each model in five evaluation metrics for 44 cohorts where a lower $AvgR_{Y_q}(m)$ indicates better performance. The ResE-BiLSTM model significantly outperforms other models, obtaining the top ranking in 37 of 44 cohorts. In contrast, although other models perform well on specific cohorts, their overall rankings are particularly lower. Specifically, BiLSTM, GRU, and RNN have the highest average rank in two cohorts each, CNN in one, and none for LSTM.

\begin{table}[!ht]
\centering
\caption{Summary of the mean ranking $AvgR_{Y_q}(m)$ for a model $m$ within each cohort}
\footnotesize
\begin{tabular*}{\textwidth}{@{\extracolsep\fill}lcccccc}
\toprule
\textbf{Quarter} & \textbf{LSTM} & \textbf{BiLSTM} & \textbf{GRU} & \textbf{CNN} & \textbf{RNN} & \textbf{ResE-BiLSTM}  \\
\midrule
2009Q1 & 2.2 & 4.8          & 3.4          & 5.4          & 4.2          & \textbf{1}   \\
2009Q2 & 3   & 5.2          & 3.6          & 2.8          & 5.4          & \textbf{1}   \\
2009Q3 & 4.8 & 4            & 3.8          & 5.2          & 2.2          & \textbf{1}   \\
2009Q4 & 3.4 & 4.2          & \textbf{1}   & 5.6          & 2            & 4.8          \\
2010Q1 & 2.8 & 4.8          & 4.2          & 5.6          & 2.6          & \textbf{1}   \\
2010Q2 & 3.2 & 4            & 2.6          & 6            & 4.2          & \textbf{1}   \\
2010Q3 & 4   & 2.2          & 3.6          & 5.8          & 4.4          & \textbf{1}   \\
2010Q4 & 3.4 & 2.6          & 4.8          & \textbf{2.2} & 2.6          & 5.4          \\
2011Q1 & 3.4 & \textbf{1.6} & 3            & 5.4          & 3.6          & 4            \\
2011Q2 & 3.2 & 3.4          & 4            & 6            & 3.4          & \textbf{1}   \\
2011Q3 & 3.4 & 2.2          & \textbf{1.2} & 5.2          & 4.2          & 4.8          \\
2011Q4 & 2.8 & 3.8          & 4.8          & 5.2          & 2.6          & \textbf{1.6} \\
2012Q1 & 3   & 4            & 2.4          & 6            & \textbf{1.8} & 3.8          \\
2012Q2 & 4   & 6            & 4.6          & 3.4          & 2            & \textbf{1}   \\
2012Q3 & 3   & 3.6          & 3            & 5.8          & 4.6          & \textbf{1}   \\
2012Q4 & 4.2 & 2.4          & 3.2          & 6            & 4.2          & \textbf{1}   \\
2013Q1 & 3.6 & 3.8          & 2.2          & 5            & 5.4          & \textbf{1}   \\
2013Q2 & 3.4 & 2.2          & 3.8          & 5.8          & 4.2          & \textbf{1.6} \\
2013Q3 & 3   & 3.6          & 2.4          & 6            & 4.2          & \textbf{1.8} \\
2013Q4 & 3.6 & 3.2          & 4.8          & 6            & 2.2          & \textbf{1.2} \\
2014Q1 & 3.8 & 4.4          & 3.2          & 6            & 2.6          & \textbf{1}   \\
2014Q2 & 3   & 3.8          & 5            & 5.4          & 2.6          & \textbf{1.2} \\
2014Q3 & 2.8 & 2.2          & 5.2          & 5.6          & 3.8          & \textbf{1.4} \\
2014Q4 & 3.2 & 4            & 4.8          & 4.8          & 3.2          & \textbf{1}   \\
2015Q1 & 3.4 & 4.4          & 5            & 4.6          & 2.4          & \textbf{1.2} \\
2015Q2 & 2.6 & 2.4          & 4.4          & 5.6          & 5            & \textbf{1}   \\
2015Q3 & 3   & 2            & 3.8          & 6            & 5            & \textbf{1.2} \\
2015Q4 & 4   & \textbf{1.4} & 3            & 6            & 3.8          & 2.8          \\
2016Q1 & 4.6 & 2.6          & 3.8          & 6            & 2.8          & \textbf{1.2} \\
2016Q2 & 3.8 & 3.8          & 2.8          & 6            & \textbf{2}   & 2.4          \\
2016Q3 & 2.4 & 4.4          & \textbf{2.2} & 6            & 3.8          & \textbf{2.2} \\
2016Q4 & 4   & 4.6          & 2.8          & 6            & 2.4          & \textbf{1.2} \\
2017Q1 & 3.6 & 3.4          & 4.6          & 6            & 2.4          & \textbf{1}   \\
2017Q2 & 3   & 3.2          & 4            & 5.2          & 4.6          & \textbf{1}   \\
2017Q3 & 3   & 2.4          & 4.4          & 6            & 4.2          & \textbf{1}   \\
2017Q4 & 3.2 & 2.6          & 4            & 6            & 4.2          & \textbf{1}   \\
2018Q1 & 2.4 & 3.8          & 3.4          & 6            & 3.6          & \textbf{1.8} \\
2018Q2 & 2.4 & 3.4          & 4            & 6            & 4.2          & \textbf{1}   \\
2018Q3 & 3.6 & 3.8          & 3.8          & 6            & 2.2          & \textbf{1.6} \\
2018Q4 & 4   & 3.6          & 3.8          & 6            & 2.6          & \textbf{1}   \\
2019Q1 & 3   & 4.2          & 4            & 6            & 2            & \textbf{1.8} \\
2019Q2 & 3.2 & 3.8          & 4.8          & 5.2          & 3            & \textbf{1}   \\
2019Q3 & 3.6 & 2.4          & 4.2          & 6            & 3            & \textbf{1.8} \\
2019Q4 & 3   & 4.4          & 3.8          & 6            & 2.8          & \textbf{1} 	\\
\bottomrule
\end{tabular*}
\label{table2}
\end{table}

\subsubsection{Ranking Performance Grouped by Year}\label{subsubsec4-1-2}

Cohorts within the same year, while independently collected, can be effectively grouped by year for model performance evaluation. This approach is valid since all four cohorts which from the same year are likely affected by similar social and market conditions. Factors such as macroeconomic trends, policy shifts, and industry-specific cycles may similarly influence data across these cohorts. By analyzing data from one year collectively, this study gains a more complete assessment of model performance stability throughout the entire year, rather than examining each quarter separately.

Using the year as a grouping unit helps mitigate the effects of seasonal variations, unexpected events, and short-term economic changes that could impact the independence of the cohorts, thus improving the robustness and generalizability of the model analysis. In finance, model stability and adaptability across years are crucial due to the significant fluctuations in financial markets and economic activities. The aggregation of yearly data avoids overemphasis on fluctuations in the single quarter, offering a more suitable evaluation of the performance of the model. This approach supports a more comprehensive performance assessment, reducing the influence of individual quarter volatility.

The average ranking method facilitates cohort grouping according to particular criteria, followed by intra-group ranking. Specifically, the four quarterly cohorts per year are grouped, and 24 results (4 cohorts $\times$ 6 models) are ranked for each metric per group. The annual ranking for each model is determined by averaging its quarterly rankings, with the annual average ranking for a model $m$ in year $Y$ defined as follows:
\begin{equation}
AvgR_{Y}(m) = \frac{1}{5 \times 4}\sum_{q=1}^4 \big( AccR_{Y}(m,q) + PreR_{Y}(m,q) + RecR_{Y}(m,q) + F1R_{Y}(m,q) + AUCR_{Y}(m,q) \big)
\end{equation}

\normalsize

\noindent where $AccR_{Y}(m,q), PreR_{Y}(m,q), RecR_{Y}(m,q), F1R_{Y}(m,q),$ and $AUCR_{Y}(m,q)$ denote the model rankings $m$ during the quarter $q$ in year $Y$, based on accuracy, precision, recall, F1, and AUC, respectively.

\begin{table}[!ht]
\centering
\caption{Summary of the annual average ranking for different models based on five different evaluation metrics}
\begin{tabular}{ccccccc}
\toprule
\textbf{Year} & \textbf{LSTM} & \textbf{BiLSTM} & \textbf{GRU} & \textbf{CNN} & \textbf{RNN} & \textbf{ResE-BiLSTM}  \\
\midrule
2009 & 12.10  & 14.10      & 11.65          & 17.30  & 12.40      & \textbf{7.45} \\
2010 & 12.15 & 11.80      & 12.70           & 15.70  & 12.35     & \textbf{10.30} \\
2011 & 10.75 & \textbf{9.55} & 10.55 & 20.20  & 12.25     & 11.65         \\
2012 & 12.50  & 12.80      & 12.10           & 16.45 & 11.80      & \textbf{9.35} \\
2013 & 12.15 & 11.40      & 11.80           & 19.40  & 13.20      & \textbf{7.05} \\
2014 & 11.85 & 12.40      & 13.95          & 18.55 & 11.40      & \textbf{6.85} \\
2015 & 11.30  & 9.95      & 12.75          & 20.65 & 12.90      & \textbf{7.45} \\
2016 & 12.60  & 12.70      & 10.70           & 20.25 & 10.65     & \textbf{8.05} \\
2017 & 11.05 & 10.75     & 13.70           & 21.75 & 12.60      & \textbf{5.15} \\
2018 & 11.45 & 11.80      & 12.20           & 21.60  & 11.60      & \textbf{6.35} \\
2019 & 11.10  & 12.45     & 12.80           & 20.85 & 10.15     & \textbf{7.65} 	\\
\bottomrule
\end{tabular}
\label{table3}
\end{table}

Table \ref{table3} presents the relative performance ranking of six models in different years grouped from the 44 cohorts. The results reveal that ResE-BiLSTM outperformed the other models in 10 of 11 years, accounting for 90.91\% of the total years. In contrast, while models such as BiLSTM displayed strong performance in some individual years, they generally exhibited lower performance compared to ResE-BiLSTM. This highlights the good consistency of the proposed ResE-BiLSTM model in delivering high performance in different annual cohorts.

\subsection{Ablation Study}

An ablation study was conducted to evaluate the behavior of ResE-BiLSTM by excluding specific components. Four model variations were created: M1 omits the residual connection mechanism, M2 omits the feedforward network, M3 omits the Residual-enhanced Encoder, and M4 removes the bidirectional feature of the BiLSTM. Data were grouped by merging three-year periods into cohorts, minimizing previous partition bias and ensuring generalizability of the results. Table \ref{table 4} concisely presents the results of the ablation study, demonstrating that all the variations of the model underperformed compared to the ResE-BiLSTM model.

\begin{table}[!h]
\centering
\caption{Overview evaluation of ablation study performance with proposed ResE-BiLSTM, E-BiLSTM, A-BiLSTM, BiLSTM, and LSTM}
\footnotesize
\begin{tabular}{llccccc}
\toprule
\textbf{Cohort} & \textbf{Metrics} & \textbf{ResE-BiLSTM}& \textbf{E-BiLSTM} & \textbf{A-BiLSTM } & \textbf{BiLSTM} & \textbf{LSTM} \\
& &  & \textbf{(M1)} & \textbf{(M2)} & \textbf{(M3)} & \textbf{(M4)} \\
\midrule
\multirow{5}*{200920102011}	&	Accuracy          & 0.9283 & 0.9151 & 0.7514 & 0.9121 & 0.9040 \\
&	Precision         & 0.9614 & 0.9493 & 0.9451 & 0.9467 & 0.9534 \\
&	Recall            & 0.8917 & 0.8670 & 0.5347 & 0.8734 & 0.8497 \\
&	F1                & 0.9252 & 0.9063 & 0.6812 & 0.9085 & 0.8984 \\
&	AUC               & 0.9709 & 0.9618 & 0.8702 & 0.9614 & 0.9594 \\
\midrule
\multirow{5}*{201220132014}	&	Accuracy          & 0.9311 & 0.9184 & 0.7460 & 0.9086 & 0.9079 \\
&	Precision         & 0.9317 & 0.9191 & 0.7421 & 0.8930 & 0.8957 \\
&	Recall            & 0.9404 & 0.9267 & 0.7750 & 0.9286 & 0.9234 \\
&	F1                & 0.9360 & 0.9229 & 0.7535 & 0.9104 & 0.9093 \\
&	AUC               & 0.9724 & 0.9612 & 0.8475 & 0.9577 & 0.9556 \\
\midrule
\multirow{5}*{201520162017}	&	Accuracy          & 0.9203 & 0.9050 & 0.7047 & 0.8933 & 0.8882 \\
&	Precision         & 0.8945 & 0.8843 & 0.6839 & 0.8811 & 0.8696 \\
&	Recall            & 0.9312 & 0.9184 & 0.7763 & 0.9094 & 0.9132 \\
&	F1                & 0.9125 & 0.9010 & 0.7241 & 0.8950 & 0.8909 \\
&	AUC               & 0.9678 & 0.9572 & 0.8101 & 0.9561 & 0.9549 \\
\midrule
\multirow{5}*{201820192020}	&	Accuracy          & 0.9331 & 0.9196 & 0.7950 & 0.9154 & 0.9120 \\
&	Precision         & 0.9791 & 0.9687 & 0.9599 & 0.9579 & 0.9579 \\
&	Recall            & 0.8671 & 0.8496 & 0.6967 & 0.8325 & 0.8257 \\
&	F1                & 0.9197 & 0.9052 & 0.8074 & 0.8908 & 0.8869 \\
&	AUC               & 0.9736 & 0.9619 & 0.9059 & 0.9593 & 0.9599 \\
\bottomrule
\end{tabular}
\label{table 4}
\end{table}

The ResE-BiLSTM model consistently achieves an accuracy of over 92\% across all cohorts. In contrast, E-BiLSTM (M1) shows slightly lower performance, indicating that removing the residual connections has some impact on overall performance, but it is not a decisive factor. A-BiLSTM (M2) exhibits the most significant performance drop, suggesting that the feedforward neural network (FNN) plays a more critical role in enhancing the model’s predictive capability. Although the M2 model incorporates an attention mechanism on BiLSTM, the absence of the FNN support leads to the attention output failing to effectively convert into discriminative features. Instead, it may increase the focus on noise or the majority class, resulting in worse performance compared to the basic BiLSTM and LSTM models. This phenomenon emphasizes the importance of the collaborative relationship between modules in this task.

Moreover, ResE-BiLSTM demonstrates excellent precision, recall, and F1 scores across all cohorts, validating the effectiveness of its structural design. E-BiLSTM (M1) shows a performance decline after the removal of residual connections, especially in recall, indicating that residual connections play a significant role in capturing deep temporal information and improving the recognition of the minority class. In contrast, A-BiLSTM (M2) experiences a more drastic performance drop after the removal of the feedforward neural network (FNN), with an average recall decrease of 23.48\% across the four cohorts, highlighting the critical importance of FNN in enhancing feature discriminability. Although BiLSTM and LSTM, which do not incorporate residual or feedforward structures, show relatively stable performance, they consistently fall short of ResE-BiLSTM in terms of all evaluation metrics.

Overall, the ablation study results indicate that each key component in the ResE-BiLSTM structure plays an irreplaceable role in model performance. Removing any of these modules leads to performance degradation across various dimensions, providing crucial insights for structural optimization in future model design.

\subsection{Interpretability Performance Analysis}

\subsubsection{Barplot Analysis}

Figures \ref{fig:figure1f} to \ref{fig:figure1b} in the appendix show SHAP barplots for the proposed ResE-BiLSTM and five baseline models, ranked in 238 features. These barplots are derived from the third cohort in the ablation study, covering years 2015 to 2017. The barplots reveal that the models prioritize different features with varying emphasis on their temporal order. Table \ref{table:appears} provides statistics for the top 50 ranked features, revealing that 14 features appear consistently each month. In addition, the findings indicate variations in feature emphasis in all six models, which explain the differences in their contributions.

\begin{table}[!h]
\centering
\caption{The number of months each feature appears in the top 50 feature importance rankings (up to a maximum of 14).}
\label{table:appears}
\footnotesize
\begin{tabular}{lcccccc}
\toprule
\textbf{Feature} &	\textbf{ResE-BiLSTM}	&	\textbf{BiLSTM}	&	\textbf{LSTM}	&	\textbf{GRU}	&	\textbf{RNN}	&	\textbf{CNN}	\\ \midrule
Interest Bearing UPB-Delta	&	14	&	14	&	14	&	14	&	14	&	14	\\
Current Actual UPB-Delta	&	14	&	14	&	14	&	14	&	14	&	14	\\
Estimated Loan to Value (ELTV)	&	12	&	11	&	14	&	14	&	11	&	14	\\
Borrower Assistance Status Code\_F	&	3	&	3	&	4	&	3	&	3	&	-	\\
Delinquency Due To Disaster\_Y	&	4	&	3	&	3	&	2	&	3	&	-	\\
Current Deferred UPB	&	3	&	3	&	-	&	3	&	4	&	8	\\
Delinquency Due To Disaster\_NAN	&	-	&	1	&	-	&	-	&	1	&	-	\\
Borrower Assistance Status Code\_NAN	&	-	&	1	&	-	&	-	&	-	&	-	\\
Current Interest Rate	&	-	&	-	&	1	&	-	&	-	&	-	\\
\bottomrule
\end{tabular}
\end{table}

For the ResE-BiLSTM model, six key features consistently rank among the top-50 over 14 months. Features such as Interest Bearing UPB-Delta, Current Actual UPB-Delta, and Estimated Loan to Value (ELTV) were significant for 14 months, 14 months and 12 months, respectively, making up 80\% of these top features, with no direct relation between feature importance and time. In contrast, the BiLSTM model identifies eight features in the top 50, with Interest Bearing UPB-Delta prominent for 14 months. Unlike ResE-BiLSTM, BiLSTM ranks feature importance chronologically, generally decreasing from recent to past months, with minor fluctuations in some months.

In the LSTM model, six features are most prominent, with Interest Bearing UPB-Delta, Current Actual UPB-Delta, and ELTV consistently appearing over 14 months, making up 84\% of the top 50 features. Unlike ResE-BiLSTM, the Current Actual UPB-Delta was identified as the most significant feature.

The GRU model highlights six features, similar to the ResE-BiLSTM model, where the importance of the feature is not linearly related to the time order in both models. However, the GRU's ranking of feature importance over time is more unpredictable and lacks a consistent pattern. Moreover, the GRU prioritizes Current Actual UPB-Delta over Interest Bearing UPB-Delta.

\begin{figure}[!hp]
\centering
    \subfloat[]{\includegraphics[width=0.85\linewidth]{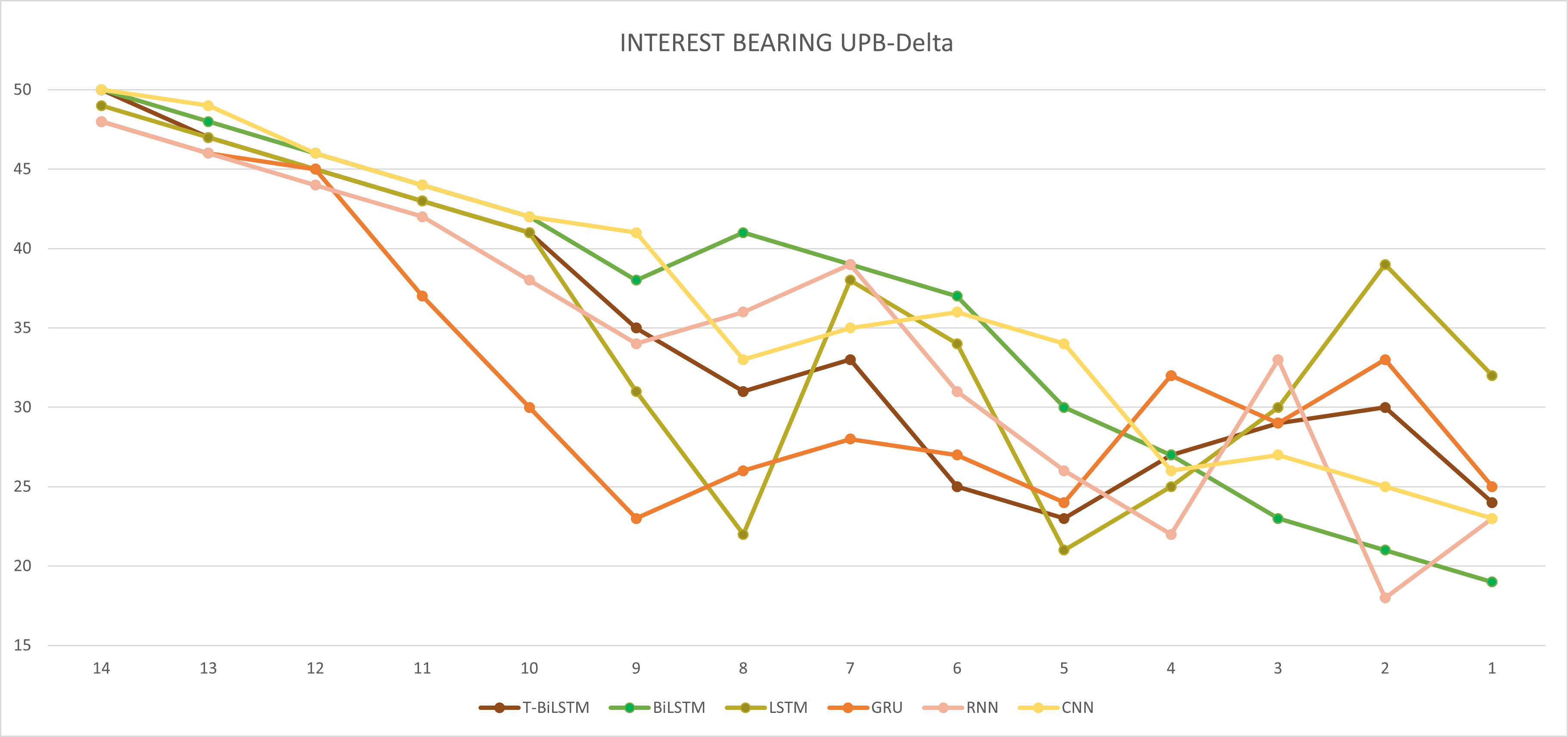} \label{fig:feature1}} \\
    \subfloat[]{\includegraphics[width=0.85\linewidth]{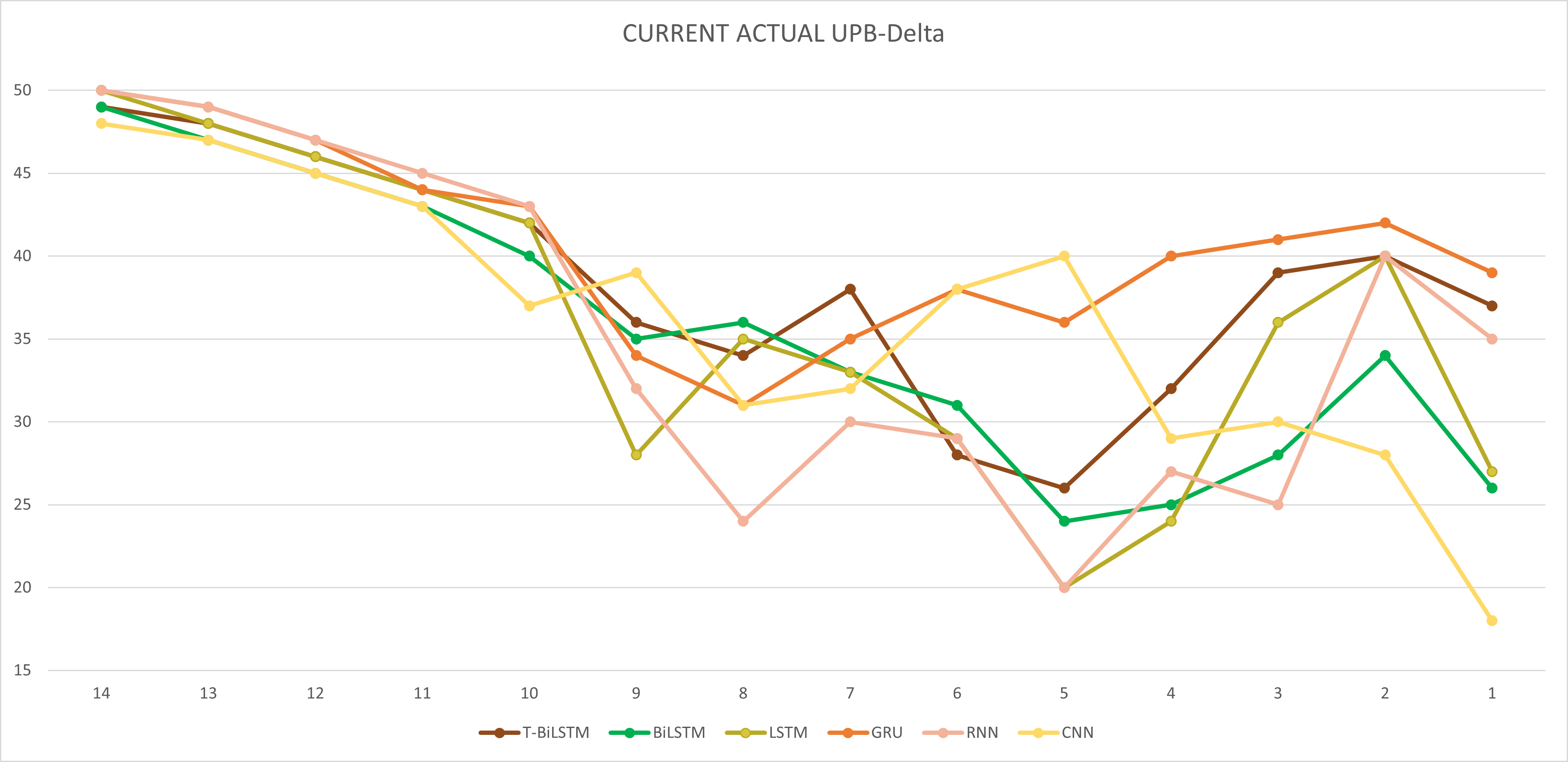} \label{fig:feature2}} \\
    \subfloat[]{\includegraphics[width=0.85\linewidth]{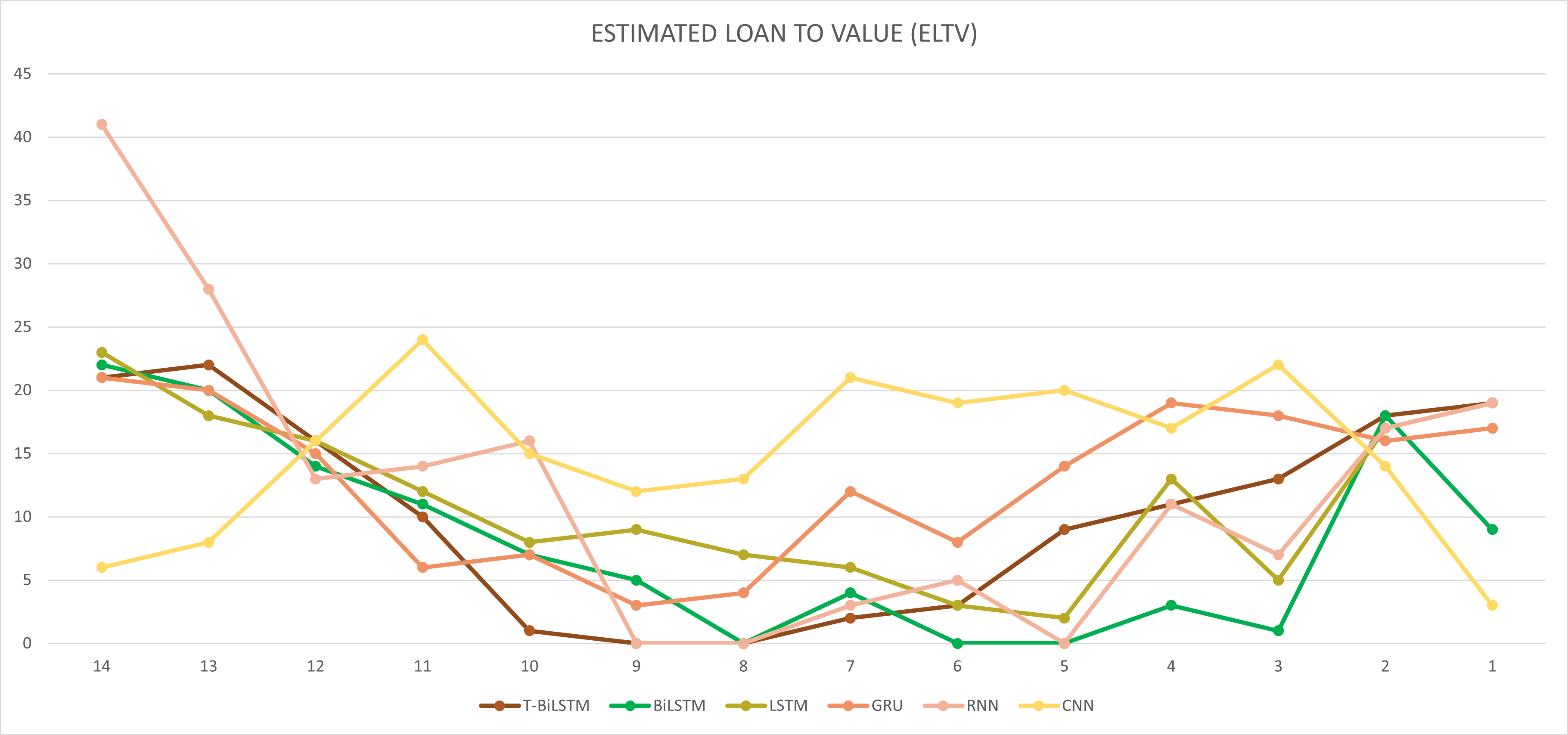} \label{fig:feature3}}
    \caption{Line charts depict the feature importance using SHAP over time for the top three features: (a) Interest Bearing UPB-Delta, (b) Current Actual UPB-Delta, and (c) ELTV. The horizontal axis denotes the time sequence, while the vertical axis shows the feature importance in the top 50 rankings, where 50 signifies the most important feature.}
\end{figure}

The results of the RNN model are the same as those of the GRU model, with the current actual UPB-Delta as the key feature. However, the ranking of feature importance throughout the sequence varies from the GRU model, showing less regularity. In contrast, the CNN model concentrates on four features, highlighting Interest Bearing UPB-Delta as most significant. For all six models, Interest Bearing UPB-Delta, Current Actual UPB-Delta, and ELTV are the most significant features. Line charts illustrating the correlation between their chronological sequence and importance are presented in Figures \ref{fig:feature1}, \ref{fig:feature2} and \ref{fig:feature3} for further analysis.

For Interest Bearing UPB-Delta, the ResE-BiLSTM and GRU models indicate that data from both distant and recent times are important for predictions, whereas intermediate periods are less significant. In contrast, the BiLSTM and CNN models show a nearly linear decrease in feature importance from recent to past data. In contrast, the LSTM and RNN models demonstrate a variable pattern without consistent changes in the importance of the features.

For Current Actual UPB-Delta, all six models show a double-peak pattern in feature importance over time, initially decreasing from recent to distant points, then rising and falling again. This pattern suggests that both recent and distant data may contain meaningful signals. Interestingly, in the GRU and CNN models, feature importance starts to increase at the 8-month (6 months before the most recent point), while in other models, this rise begins at the 5-month (9 months earlier).

For ELTV, the general importance of the features over the 14 months is less than the previous two features, usually ranking in the bottom half of the top 50 with mild month-to-month variation. Besides the CNN model, the other five models show lower feature importance in the mid-periods, increasing at both timeline extremes. Moreover, the evaluation of the importance of the features in different models reveals similar trends in similar time frames.

\subsubsection{SHAP Summary Plot Analysis}

The SHAP summary plot (Figure \ref{fig:figure2f} to \ref{fig:figure2b} in Appendix) depicts the influence and significance of each feature on the model outcome, both positively and negatively. The distribution of positions and colors reveals how variations in feature values affect prediction results. Specifically, each dot represents a sample, with the vertical stacking of dots indicating sample density. The horizontal position corresponds to the SHAP value of the feature, which reflects the magnitude and direction of the feature's contribution to the model prediction. The vertical axis ranks features based on the sum of SHAP values across all samples, following the same order as in the bar plot. A SHAP value positioned further to the right indicates a stronger positive contribution to the prediction, while values further to the left indicate a stronger negative contribution. A larger horizontal spread signifies that the range of the feature’s values has a more significant impact on the model's predictions. The color coding represents the magnitude of the feature values, with red indicating higher values and blue indicating lower values.

The findings demonstrated that the lower value of Interest Bearing UPB-Delta and Current Actual UPB-Delta significantly improve the model computation, while the lower value of ELTV negatively impacts it, as the blue dots are concentrated on the negative
side (left). Furthermore, five models (besides the CNN model) displayed varying patterns, suggesting that these features may impact positively or negatively based on the model. Moreover, the CNN model uniquely assessed the significance of these two features. The CNN model’s differing assessment of the importance of these two key features may explain why its performance is inferior to that of the other models.

The six models also differ in how they assess the impact of ELTV on prediction outcomes. The ResE-BiLSTM model indicates that ELTV has a dual effect, sometimes exerting minimal influence. For BiLSTM and LSTM models, ELTV's impact fluctuates, potentially due to the sensitivity of the time series data or shifts in its relationship with the prediction target over time. For features of lesser importance, such as Delinquency due to Disaster\_Y, all models similarly evaluate their contribution. Likewise, Current Deferred UPB exhibited both positive and negative impacts across models. Thus, these features are not the main factors differentiating model performance.

In summary, the different models show variation in their assessments of feature importance, and the impact of features on prediction results changes over time. The differences in model responses and the assessment of feature importance reveal the models' varying abilities to capture feature complexity and time-series characteristics.

\section{Conclusion}

This study addresses loan default prediction by introducing the ResE-BiLSTM model, which consistently outperforms baseline models in accuracy, precision, recall, F1 score, and AUC for most cohorts. The ResE-BiLSTM performance highlights the promise of multi-layered models in capturing complex data patterns and improving prediction accuracy. In addition, the study highlights the efficacy of Residual-enhanced Encoder and BiLSTM elements in anomaly detection. The interpretability analysis examines the significance of the features and variations in the importance of the features in the models in time series. These results shed light on the inner workings of the models, which aids in future optimization. Future work will focus on refine these model components and investigate more efficient anomaly detection techniques to assist financial institutions in identifying high-risk borrowers, minimizing non-performing loans, and enhancing asset quality and financial stability.

\newpage
\section{List of abbreviations}
\begin{table}[!htbp]
\centering
\renewcommand{\arraystretch}{1.5} 
\begin{tabular*}{0.7\textwidth}{@{\extracolsep\fill}cc}
\toprule
\textbf{Abbreviation} & \textbf{Definition} \\ 
\midrule
LSTM  & long-short-term memory \\
BiLSTM  & bidirectional long-short-term memory \\
XAI & Explainable AI \\
OOS & out-of-sample \\
SHAP & SHapley Additive exPlanations \\
CNN & convolutional neural networks \\
RNN & recurrent neural networks \\
LIME & Local Interpretable Model-Agnostic Explanation \\
ELTV & estimated loan to value \\
AUC & area under the ROC curve \\
\bottomrule
\end{tabular*}
\caption*{} 
\end{table}










\bibliography{ResEBiLSTMref.bib}

\newpage
\begin{appendices}

\section*{Appendix}\label{secA1}
\setcounter{table}{0}
\renewcommand{\thetable}{A\arabic{table}}
\setcounter{figure}{0}
\renewcommand{\thefigure}{A\arabic{figure}}

\begin{table}[h]
\centering
\caption{Accuracy of 6 models on 44 cohorts of the Freddie Mac dataset based on average results from 10 trials.}
\label{acc}
\footnotesize
\begin{tabular*}{\textwidth}{@{\extracolsep\fill}lcccccc}
\toprule
\multirow{2}*{\textbf{Cohort}} & \textbf{LSTM} & \textbf{BiLSTM} & \textbf{GRU} & \textbf{CNN} & \textbf{RNN} & \multirow{2}*{\textbf{ResE-BiLSTM}} \\
& \citeyear{2021_9} & \citeyear{2019_8} & \citeyear{2021_13} & \citeyear{Kakkar_2023}& \citeyear{RNN}\\
\midrule
2009Q1 & 0.918 & 0.913          & 0.914          & 0.892          & 0.914          & \textbf{0.923} \\
2009Q2 & 0.895 & 0.880          & 0.887          & 0.897          & 0.883          & \textbf{0.921} \\
2009Q3 & 0.914 & 0.914          & 0.915          & 0.911          & 0.919          & \textbf{0.924} \\
2009Q4 & 0.926 & 0.924          & \textbf{0.945} & 0.906          & 0.937          & 0.924          \\
2010Q1 & 0.939 & 0.935          & 0.938          & 0.935          & 0.940          & \textbf{0.946} \\
2010Q2 & 0.901 & 0.900          & 0.903          & 0.887          & 0.901          & \textbf{0.913} \\
2010Q3 & 0.909 & 0.919          & 0.913          & 0.893          & 0.911          & \textbf{0.922} \\
2010Q4 & 0.899 & 0.900          & 0.896          & \textbf{0.911} & 0.910          & 0.890          \\
2011Q1 & 0.925 & \textbf{0.933} & 0.927          & 0.904          & 0.924          & 0.903          \\
2011Q2 & 0.922 & 0.921          & 0.919          & 0.895          & 0.920          & \textbf{0.923} \\
2011Q3 & 0.922 & 0.929          & \textbf{0.935} & 0.896          & 0.910          & 0.906          \\
2011Q4 & 0.920 & 0.916          & 0.912          & 0.886          & 0.920          & \textbf{0.925} \\
2012Q1 & 0.904 & 0.890          & 0.905          & 0.875          & \textbf{0.919} & 0.894          \\
2012Q2 & 0.830 & 0.817          & 0.827          & 0.848          & 0.855          & \textbf{0.863} \\
2012Q3 & 0.911 & 0.910          & 0.911          & 0.863          & 0.902          & \textbf{0.913} \\
2012Q4 & 0.948 & 0.953          & 0.949          & 0.935          & 0.947          & \textbf{0.954} \\
2013Q1 & 0.878 & 0.878          & 0.884          & 0.868          & 0.865          & \textbf{0.916} \\
2013Q2 & 0.926 & 0.931          & 0.923          & 0.889          & 0.913          & \textbf{0.932} \\
2013Q3 & 0.893 & 0.890          & 0.897          & 0.857          & 0.885          & \textbf{0.909} \\
2013Q4 & 0.914 & 0.914          & 0.910          & 0.887          & 0.924          & \textbf{0.930} \\
2014Q1 & 0.921 & 0.915          & 0.924          & 0.886          & 0.927          & \textbf{0.931} \\
2014Q2 & 0.918 & 0.917          & 0.912          & 0.885          & 0.918          & \textbf{0.924} \\
2014Q3 & 0.931 & 0.933          & 0.924          & 0.919          & 0.927          & \textbf{0.935} \\
2014Q4 & 0.875 & 0.865          & 0.862          & 0.869          & 0.884          & \textbf{0.891} \\
2015Q1 & 0.896 & 0.885          & 0.879          & 0.892          & 0.910          & \textbf{0.913} \\
2015Q2 & 0.911 & 0.912          & 0.904          & 0.890          & 0.903          & \textbf{0.916} \\
2015Q3 & 0.926 & 0.930          & 0.924          & 0.894          & 0.914          & \textbf{0.931} \\
2015Q4 & 0.927 & \textbf{0.936} & 0.928          & 0.884          & 0.928          & 0.908          \\
2016Q1 & 0.913 & 0.920          & 0.916          & 0.886          & 0.919          & \textbf{0.927} \\
2016Q2 & 0.908 & 0.909          & 0.913          & 0.886          & 0.913          & \textbf{0.923} \\
2016Q3 & 0.912 & 0.901          & 0.909          & 0.879          & 0.907          & \textbf{0.915} \\
2016Q4 & 0.930 & 0.925          & 0.934          & 0.913          & 0.940          & \textbf{0.941} \\
2017Q1 & 0.927 & 0.925          & 0.923          & 0.902          & 0.929          & \textbf{0.933} \\
2017Q2 & 0.921 & 0.918          & 0.910          & 0.897          & 0.909          & \textbf{0.930} \\
2017Q3 & 0.916 & 0.916          & 0.914          & 0.883          & 0.914          & \textbf{0.923} \\
2017Q4 & 0.925 & 0.925          & 0.923          & 0.901          & 0.923          & \textbf{0.930} \\
2018Q1 & 0.937 & 0.936          & 0.937          & 0.915          & 0.935          & \textbf{0.939} \\
2018Q2 & 0.928 & 0.926          & 0.925          & 0.899          & 0.923          & \textbf{0.934} \\
2018Q3 & 0.930 & 0.930          & 0.930          & 0.911          & 0.932          & \textbf{0.935} \\
2018Q4 & 0.919 & 0.923          & 0.922          & 0.900          & 0.926          & \textbf{0.927} \\
2019Q1 & 0.926 & 0.924          & 0.923          & 0.907          & 0.927          & \textbf{0.930} \\
2019Q2 & 0.932 & 0.930          & 0.927          & 0.923          & 0.937          & \textbf{0.942} \\
2019Q3 & 0.944 & 0.949          & 0.943          & 0.921          & 0.946          & \textbf{0.951} \\
2019Q4 & 0.951 & 0.941          & 0.947          & 0.903          & 0.953          & \textbf{0.955}   \\
\bottomrule
\label{tab:A1}
\end{tabular*}
\end{table}

\begin{table}[!h]
\centering
\caption{Precision of 6 models on 44 cohorts of the Freddie Mac dataset based on average results from 10 trials.}
\label{pre}
\footnotesize
\begin{tabular*}{\textwidth}{@{\extracolsep\fill}lcccccc}
\toprule
\multirow{2}*{\textbf{Cohort}} & \textbf{LSTM} & \textbf{BiLSTM} & \textbf{GRU} & \textbf{CNN} & \textbf{RNN} & \multirow{2}*{\textbf{ResE-BiLSTM}} \\
& \citeyear{2021_9} & \citeyear{2019_8} & \citeyear{2021_13} & \citeyear{Kakkar_2023}& \citeyear{RNN}\\
\midrule
2009Q1 & 0.950          & 0.949          & 0.945          & 0.950          & 0.950          & \textbf{0.951} \\
2009Q2 & 0.897          & 0.881          & 0.892          & 0.901          & 0.885          & \textbf{0.953} \\
2009Q3 & 0.921          & 0.919          & 0.922          & 0.925          & 0.922          & \textbf{0.927} \\
2009Q4 & 0.964          & 0.951          & \textbf{0.965} & 0.959          & 0.964          & 0.956          \\
2010Q1 & 0.978          & 0.980          & 0.977          & 0.974          & 0.982          & \textbf{0.983} \\
2010Q2 & 0.895          & 0.893          & 0.900          & 0.888          & 0.904          & \textbf{0.915} \\
2010Q3 & 0.894          & 0.919          & 0.911          & 0.896          & 0.917          & \textbf{0.920} \\
2010Q4 & 0.923          & 0.921          & 0.915          & \textbf{0.970} & 0.954          & 0.919          \\
2011Q1 & 0.929          & 0.928          & \textbf{0.935} & 0.915          & 0.915          & 0.862          \\
2011Q2 & 0.969          & 0.973          & 0.976          & 0.934          & 0.985          & \textbf{0.988} \\
2011Q3 & 0.941          & 0.935          & \textbf{0.954} & 0.949          & 0.908          & 0.902          \\
2011Q4 & 0.940          & 0.925          & 0.919          & 0.945          & \textbf{0.956} & 0.935          \\
2012Q1 & 0.893          & 0.863          & 0.893          & 0.859          & \textbf{0.923} & 0.863          \\
2012Q2 & 0.785          & 0.770          & 0.783          & 0.807          & 0.816          & \textbf{0.828} \\
2012Q3 & 0.920          & 0.919          & 0.916          & 0.851          & 0.932          & \textbf{0.933} \\
2012Q4 & 0.985          & 0.985          & 0.983          & 0.971          & 0.979          & \textbf{0.990} \\
2013Q1 & 0.854          & 0.852          & 0.862          & 0.879          & 0.834          & \textbf{0.911} \\
2013Q2 & 0.981          & \textbf{0.987} & 0.978          & 0.922          & 0.981          & 0.980          \\
2013Q3 & 0.869          & 0.865          & 0.877          & 0.829          & 0.855          & \textbf{0.903} \\
2013Q4 & 0.919          & 0.913          & 0.918          & 0.901          & \textbf{0.937} & 0.920          \\
2014Q1 & 0.931          & 0.917          & 0.939          & 0.883          & 0.946          & \textbf{0.948} \\
2014Q2 & 0.930          & 0.929          & 0.925          & 0.931          & \textbf{0.944} & 0.936          \\
2014Q3 & \textbf{0.952} & 0.950          & 0.939          & 0.945          & 0.945          & 0.949          \\
2014Q4 & 0.850          & 0.834          & 0.827          & 0.872          & 0.870          & \textbf{0.874} \\
2015Q1 & 0.875          & 0.854          & 0.847          & 0.892          & \textbf{0.905} & 0.901          \\
2015Q2 & 0.925          & 0.923          & 0.919          & 0.922          & 0.913          & \textbf{0.936} \\
2015Q3 & 0.932          & \textbf{0.938} & 0.926          & 0.890          & 0.912          & 0.936          \\
2015Q4 & 0.931          & \textbf{0.946} & 0.935          & 0.910          & 0.940          & 0.941          \\
2016Q1 & 0.928          & 0.935          & 0.928          & 0.905          & \textbf{0.961} & 0.941          \\
2016Q2 & 0.916          & 0.928          & 0.927          & 0.887          & 0.921          & \textbf{0.941} \\
2016Q3 & 0.915          & 0.893          & 0.907          & 0.881          & 0.904          & \textbf{0.929} \\
2016Q4 & 0.950          & 0.939          & 0.954          & 0.933          & \textbf{0.971} & 0.958          \\
2017Q1 & 0.955          & 0.941          & 0.938          & 0.934          & 0.955          & \textbf{0.957} \\
2017Q2 & 0.926          & 0.913          & 0.898          & 0.927          & 0.894          & \textbf{0.939} \\
2017Q3 & 0.943          & 0.944          & 0.941          & 0.925          & 0.942          & \textbf{0.947} \\
2017Q4 & 0.956          & 0.957          & 0.950          & 0.927          & 0.959          & \textbf{0.961} \\
2018Q1 & 0.977          & 0.975          & 0.976          & 0.943          & \textbf{0.979} & 0.963          \\
2018Q2 & 0.958          & 0.953          & 0.957          & 0.914          & 0.955          & \textbf{0.963} \\
2018Q3 & 0.968          & 0.963          & 0.965          & 0.939          & \textbf{0.970} & 0.964          \\
2018Q4 & 0.915          & 0.927          & 0.923          & 0.903          & 0.938          & \textbf{0.940} \\
2019Q1 & 0.957          & 0.956          & \textbf{0.958} & 0.932          & 0.958          & 0.947          \\
2019Q2 & 0.911          & 0.909          & 0.904          & 0.926          & 0.922          & \textbf{0.931} \\
2019Q3 & 0.940          & 0.953          & 0.940          & 0.930          & 0.949          & \textbf{0.967} \\
2019Q4 & 0.940          & 0.918          & 0.929          & 0.897          & 0.949          & \textbf{0.953}     \\
\bottomrule
\label{tab:A2}
\end{tabular*}
\end{table}

\begin{table}[!h]
\centering
\caption{Recall rate of 6 models on 44 cohorts of the Freddie Mac dataset based on average results from 10 trials.}
\label{rec}
\footnotesize
\begin{tabular*}{\textwidth}{@{\extracolsep\fill}lcccccc}
\toprule
\multirow{2}*{\textbf{Cohort}} & \textbf{LSTM} & \textbf{BiLSTM} & \textbf{GRU} & \textbf{CNN} & \textbf{RNN} & \multirow{2}*{\textbf{ResE-BiLSTM}} \\
& \citeyear{2021_9} & \citeyear{2019_8} & \citeyear{2021_13} & \citeyear{Kakkar_2023}& \citeyear{RNN}\\
\midrule
2009Q1 & \textbf{0.883} & 0.872          & 0.880          & 0.829 & 0.874          & \textbf{0.883} \\
2009Q2 & 0.892          & 0.880          & 0.881          & 0.892 & 0.879          & \textbf{0.896} \\
2009Q3 & 0.905          & 0.909          & 0.907          & 0.894 & 0.915          & \textbf{0.921} \\
2009Q4 & 0.886          & 0.894          & \textbf{0.923} & 0.849 & 0.908          & 0.889          \\
2010Q1 & 0.898          & 0.889          & 0.897          & 0.894 & 0.896          & \textbf{0.908} \\
2010Q2 & 0.910          & 0.909          & 0.906          & 0.887 & 0.897          & \textbf{0.911} \\
2010Q3 & 0.929          & 0.918          & 0.916          & 0.891 & 0.905          & \textbf{0.934} \\
2010Q4 & 0.870          & \textbf{0.876} & 0.873          & 0.847 & 0.861          & 0.856          \\
2011Q1 & 0.921          & 0.938          & 0.918          & 0.892 & 0.936          & \textbf{0.961} \\
2011Q2 & 0.873          & 0.866          & 0.860          & 0.851 & 0.854          & \textbf{0.874} \\
2011Q3 & 0.902          & \textbf{0.923} & 0.915          & 0.837 & 0.914          & 0.912          \\
2011Q4 & 0.897          & 0.907          & 0.905          & 0.820 & 0.880          & \textbf{0.914} \\
2012Q1 & 0.917          & 0.927          & 0.922          & 0.900 & 0.913          & \textbf{0.937} \\
2012Q2 & 0.914          & 0.907          & 0.908          & 0.916 & 0.919          & \textbf{0.922} \\
2012Q3 & 0.900          & 0.899          & 0.905          & 0.881 & 0.868          & \textbf{0.912} \\
2012Q4 & 0.910          & 0.920          & 0.914          & 0.897 & 0.913          & \textbf{0.928} \\
2013Q1 & 0.912          & 0.914          & 0.915          & 0.854 & 0.912          & \textbf{0.922} \\
2013Q2 & 0.869          & 0.873          & 0.865          & 0.851 & 0.843          & \textbf{0.883} \\
2013Q3 & 0.925          & 0.926          & 0.925          & 0.900 & \textbf{0.928} & 0.917          \\
2013Q4 & 0.909          & 0.915          & 0.900          & 0.870 & 0.908          & \textbf{0.941} \\
2014Q1 & 0.910          & 0.912          & 0.908          & 0.891 & 0.905          & \textbf{0.913} \\
2014Q2 & 0.903          & 0.904          & 0.897          & 0.831 & 0.890          & \textbf{0.910} \\
2014Q3 & 0.909          & 0.913          & 0.907          & 0.891 & 0.907          & \textbf{0.920} \\
2014Q4 & 0.911          & 0.912          & 0.917          & 0.866 & 0.902          & \textbf{0.919} \\
2015Q1 & 0.924          & 0.928          & 0.929          & 0.894 & 0.916          & \textbf{0.932} \\
2015Q2 & 0.895          & 0.898          & 0.887          & 0.853 & 0.891          & \textbf{0.899} \\
2015Q3 & 0.918          & 0.921          & 0.921          & 0.898 & 0.917          & \textbf{0.926} \\
2015Q4 & 0.923          & 0.925          & 0.921          & 0.853 & 0.915          & \textbf{0.931} \\
2016Q1 & 0.895          & 0.902          & 0.902          & 0.863 & 0.873          & \textbf{0.904} \\
2016Q2 & 0.898          & 0.888          & 0.896          & 0.886 & \textbf{0.903} & 0.890          \\
2016Q3 & 0.909          & 0.911          & \textbf{0.913} & 0.877 & 0.911          & 0.897          \\
2016Q4 & 0.909          & 0.909          & 0.913          & 0.891 & 0.906          & \textbf{0.923} \\
2017Q1 & 0.897          & 0.906          & 0.906          & 0.867 & 0.901          & \textbf{0.907} \\
2017Q2 & 0.916          & 0.924          & 0.925          & 0.864 & 0.927          & \textbf{0.930} \\
2017Q3 & 0.885          & 0.884          & 0.884          & 0.833 & 0.882          & \textbf{0.896} \\
2017Q4 & 0.891          & 0.890          & 0.893          & 0.871 & 0.885          & \textbf{0.907} \\
2018Q1 & 0.896          & 0.894          & 0.896          & 0.884 & 0.889          & \textbf{0.908} \\
2018Q2 & 0.895          & 0.896          & 0.891          & 0.882 & 0.889          & \textbf{0.903} \\
2018Q3 & 0.890          & 0.894          & 0.892          & 0.879 & 0.891          & \textbf{0.904} \\
2018Q4 & 0.923          & 0.918          & 0.920          & 0.896 & 0.913          & \textbf{0.923} \\
2019Q1 & 0.892          & 0.888          & 0.885          & 0.878 & 0.895          & \textbf{0.911} \\
2019Q2 & 0.957          & 0.957          & 0.957          & 0.920 & 0.955          & \textbf{0.958} \\
2019Q3 & \textbf{0.948} & 0.944          & 0.946          & 0.910 & 0.942          & 0.942          \\
2019Q4 & 0.964          & 0.969          & 0.968          & 0.910 & 0.957          & \textbf{0.970}  \\
\bottomrule
\label{tab:A3}
\end{tabular*}
\end{table}

\begin{table}[!h]
\centering
\caption{Binary F1 score of 6 models on 44 cohorts of the Freddie Mac dataset based on average results from 10 trials.}
\label{f1}
\footnotesize
\begin{tabular*}{\textwidth}{@{\extracolsep\fill}lcccccc}
\toprule
\multirow{2}*{\textbf{Cohort}} & \textbf{LSTM} & \textbf{BiLSTM} & \textbf{GRU} & \textbf{CNN} & \textbf{RNN} & \multirow{2}*{\textbf{ResE-BiLSTM}} \\
& \citeyear{2021_9} & \citeyear{2019_8} & \citeyear{2021_13} & \citeyear{Kakkar_2023}& \citeyear{RNN}\\
\midrule
2009Q1 & 0.915 & 0.909          & 0.911          & 0.885 & 0.910          & \textbf{0.916} \\
2009Q2 & 0.894 & 0.880          & 0.886          & 0.896 & 0.882          & \textbf{0.924} \\
2009Q3 & 0.913 & 0.914          & 0.914          & 0.909 & 0.919          & \textbf{0.924} \\
2009Q4 & 0.923 & 0.921          & \textbf{0.943} & 0.900 & 0.935          & 0.921          \\
2010Q1 & 0.936 & 0.932          & 0.935          & 0.932 & 0.937          & \textbf{0.944} \\
2010Q2 & 0.902 & 0.901          & 0.903          & 0.887 & 0.900          & \textbf{0.913} \\
2010Q3 & 0.911 & 0.918          & 0.914          & 0.893 & 0.911          & \textbf{0.927} \\
2010Q4 & 0.896 & 0.898          & 0.894          & 0.904 & \textbf{0.905} & 0.886          \\
2011Q1 & 0.925 & \textbf{0.933} & 0.926          & 0.903 & 0.925          & 0.909          \\
2011Q2 & 0.918 & 0.916          & 0.914          & 0.890 & 0.915          & \textbf{0.927} \\
2011Q3 & 0.921 & 0.929          & \textbf{0.934} & 0.889 & 0.911          & 0.907          \\
2011Q4 & 0.918 & 0.916          & 0.911          & 0.877 & 0.916          & \textbf{0.924} \\
2012Q1 & 0.905 & 0.894          & 0.907          & 0.878 & \textbf{0.918} & 0.898          \\
2012Q2 & 0.844 & 0.833          & 0.840          & 0.858 & 0.864          & \textbf{0.873} \\
2012Q3 & 0.910 & 0.909          & 0.911          & 0.865 & 0.899          & \textbf{0.922} \\
2012Q4 & 0.946 & 0.951          & 0.947          & 0.932 & 0.945          & \textbf{0.958} \\
2013Q1 & 0.882 & 0.882          & 0.887          & 0.866 & 0.871          & \textbf{0.917} \\
2013Q2 & 0.921 & 0.927          & 0.918          & 0.885 & 0.907          & \textbf{0.929} \\
2013Q3 & 0.896 & 0.894          & 0.900          & 0.863 & 0.890          & \textbf{0.910} \\
2013Q4 & 0.914 & 0.914          & 0.909          & 0.885 & 0.922          & \textbf{0.930} \\
2014Q1 & 0.920 & 0.915          & 0.923          & 0.887 & 0.925          & \textbf{0.930} \\
2014Q2 & 0.916 & 0.916          & 0.911          & 0.878 & 0.916          & \textbf{0.923} \\
2014Q3 & 0.930 & 0.932          & 0.922          & 0.917 & 0.925          & \textbf{0.934} \\
2014Q4 & 0.879 & 0.871          & 0.869          & 0.869 & 0.886          & \textbf{0.896} \\
2015Q1 & 0.899 & 0.890          & 0.885          & 0.892 & 0.910          & \textbf{0.916} \\
2015Q2 & 0.909 & 0.910          & 0.903          & 0.886 & 0.902          & \textbf{0.917} \\
2015Q3 & 0.925 & 0.929          & 0.923          & 0.894 & 0.914          & \textbf{0.931} \\
2015Q4 & 0.927 & 0.935          & 0.928          & 0.880 & 0.927          & \textbf{0.936} \\
2016Q1 & 0.911 & 0.918          & 0.914          & 0.884 & 0.915          & \textbf{0.922} \\
2016Q2 & 0.907 & 0.907          & 0.911          & 0.886 & 0.912          & \textbf{0.915} \\
2016Q3 & 0.912 & 0.902          & 0.910          & 0.879 & 0.907          & \textbf{0.913} \\
2016Q4 & 0.929 & 0.924          & 0.933          & 0.911 & 0.938          & \textbf{0.940} \\
2017Q1 & 0.925 & 0.923          & 0.922          & 0.898 & 0.927          & \textbf{0.931} \\
2017Q2 & 0.921 & 0.918          & 0.911          & 0.894 & 0.910          & \textbf{0.935} \\
2017Q3 & 0.913 & 0.913          & 0.911          & 0.876 & 0.911          & \textbf{0.921} \\
2017Q4 & 0.922 & 0.922          & 0.921          & 0.898 & 0.920          & \textbf{0.933} \\
2018Q1 & 0.934 & 0.933          & 0.934          & 0.913 & 0.931          & \textbf{0.935} \\
2018Q2 & 0.925 & 0.924          & 0.923          & 0.897 & 0.921          & \textbf{0.932} \\
2018Q3 & 0.927 & 0.927          & 0.927          & 0.908 & 0.929          & \textbf{0.933} \\
2018Q4 & 0.919 & 0.922          & 0.922          & 0.899 & 0.925          & \textbf{0.931} \\
2019Q1 & 0.923 & 0.921          & 0.920          & 0.904 & 0.925          & \textbf{0.928} \\
2019Q2 & 0.934 & 0.932          & 0.930          & 0.923 & 0.938          & \textbf{0.944} \\
2019Q3 & 0.944 & 0.948          & 0.943          & 0.920 & 0.945          & \textbf{0.954} \\
2019Q4 & 0.952 & 0.943          & 0.948          & 0.903 & 0.953          & \textbf{0.961}   	\\
\bottomrule
\label{tab:A4}
\end{tabular*}
\end{table}

\begin{table}[!h]
\centering
\caption{AUC value of 6 models on 44 cohorts of the Freddie Mac dataset based on average results from 10 trials.}
\label{auc}
\footnotesize
\begin{tabular*}{\textwidth}{@{\extracolsep\fill}lcccccc}
\toprule
\multirow{2}*{\textbf{Cohort}} & \textbf{LSTM} & \textbf{BiLSTM} & \textbf{GRU} & \textbf{CNN} & \textbf{RNN} & \multirow{2}*{\textbf{ResE-BiLSTM}} \\
& \citeyear{2021_9} & \citeyear{2019_8} & \citeyear{2021_13} & \citeyear{Kakkar_2023}& \citeyear{RNN}\\
\midrule
2009Q1 & 0.968 & 0.967          & 0.969          & 0.954          & 0.964          & \textbf{0.971} \\
2009Q2 & 0.955 & 0.956          & 0.958          & 0.952          & 0.948          & \textbf{0.969} \\
2009Q3 & 0.970 & 0.971          & 0.969          & 0.966          & 0.971          & \textbf{0.972} \\
2009Q4 & 0.977 & 0.976          & \textbf{0.980} & 0.963          & 0.979          & 0.972          \\
2010Q1 & 0.985 & 0.983          & 0.983          & 0.977          & 0.985          & \textbf{0.986} \\
2010Q2 & 0.957 & 0.958          & 0.961          & 0.946          & 0.955          & \textbf{0.963} \\
2010Q3 & 0.970 & 0.970          & 0.968          & 0.957          & 0.964          & \textbf{0.972} \\
2010Q4 & 0.963 & 0.965          & 0.952          & \textbf{0.975} & 0.957          & 0.955          \\
2011Q1 & 0.972 & \textbf{0.975} & 0.970          & 0.962          & 0.972          & 0.974          \\
2011Q2 & 0.973 & 0.973          & 0.975          & 0.964          & 0.979          & \textbf{0.981} \\
2011Q3 & 0.967 & 0.969          & \textbf{0.971} & 0.960          & 0.964          & 0.966          \\
2011Q4 & 0.968 & 0.966          & 0.965          & 0.958          & 0.971          & \textbf{0.973} \\
2012Q1 & 0.968 & 0.967          & 0.970          & 0.943          & \textbf{0.972} & 0.966          \\
2012Q2 & 0.930 & 0.925          & 0.930          & 0.928          & 0.943          & \textbf{0.945} \\
2012Q3 & 0.969 & 0.971          & 0.968          & 0.938          & 0.965          & \textbf{0.974} \\
2012Q4 & 0.975 & 0.976          & 0.979          & 0.973          & 0.979          & \textbf{0.981} \\
2013Q1 & 0.956 & 0.959          & 0.960          & 0.931          & 0.955          & \textbf{0.962} \\
2013Q2 & 0.977 & 0.977          & 0.978          & 0.959          & 0.977          & \textbf{0.979} \\
2013Q3 & 0.958 & 0.958          & 0.959          & 0.938          & 0.951          & \textbf{0.960} \\
2013Q4 & 0.968 & 0.968          & 0.967          & 0.949          & 0.972          & \textbf{0.974} \\
2014Q1 & 0.969 & 0.968          & 0.970          & 0.949          & 0.972          & \textbf{0.973} \\
2014Q2 & 0.970 & 0.970          & 0.967          & 0.948          & 0.972          & \textbf{0.977} \\
2014Q3 & 0.968 & 0.970          & 0.965          & 0.964          & 0.970          & \textbf{0.971} \\
2014Q4 & 0.952 & 0.951          & 0.946          & 0.932          & 0.949          & \textbf{0.953} \\
2015Q1 & 0.961 & 0.958          & 0.957          & 0.948          & 0.964          & \textbf{0.966} \\
2015Q2 & 0.962 & 0.962          & 0.960          & 0.949          & 0.959          & \textbf{0.963} \\
2015Q3 & 0.966 & 0.965          & 0.963          & 0.948          & 0.960          & \textbf{0.967} \\
2015Q4 & 0.971 & \textbf{0.972} & 0.971          & 0.946          & 0.970          & 0.965          \\
2016Q1 & 0.957 & 0.965          & 0.965          & 0.948          & 0.965          & \textbf{0.969} \\
2016Q2 & 0.952 & 0.950          & 0.952          & 0.944          & \textbf{0.953} & 0.949          \\
2016Q3 & 0.959 & 0.955          & \textbf{0.960} & 0.939          & 0.956          & 0.958          \\
2016Q4 & 0.966 & 0.964          & 0.967          & 0.960          & 0.971          & \textbf{0.971} \\
2017Q1 & 0.960 & 0.960          & 0.960          & 0.952          & 0.962          & \textbf{0.970} \\
2017Q2 & 0.960 & 0.962          & 0.959          & 0.945          & 0.959          & \textbf{0.963} \\
2017Q3 & 0.958 & 0.960          & 0.957          & 0.938          & 0.959          & \textbf{0.964} \\
2017Q4 & 0.963 & 0.964          & 0.962          & 0.946          & 0.961          & \textbf{0.973} \\
2018Q1 & 0.969 & 0.970          & 0.969          & 0.960          & 0.971          & \textbf{0.972} \\
2018Q2 & 0.967 & 0.966          & 0.966          & 0.953          & 0.967          & \textbf{0.969} \\
2018Q3 & 0.965 & 0.967          & 0.967          & 0.956          & 0.969          & \textbf{0.973} \\
2018Q4 & 0.970 & 0.969          & 0.969          & 0.959          & 0.971          & \textbf{0.973} \\
2019Q1 & 0.975 & 0.973          & 0.974          & 0.968          & 0.978          & \textbf{0.978} \\
2019Q2 & 0.980 & 0.980          & 0.979          & 0.972          & 0.980          & \textbf{0.983} \\
2019Q3 & 0.978 & 0.979          & 0.978          & 0.969          & 0.981          & \textbf{0.983} \\
2019Q4 & 0.984 & 0.980          & 0.982          & 0.956          & 0.983          & \textbf{0.986}	\\
\bottomrule
\label{tab:A5}
\end{tabular*}
\end{table}

\begin{figure}[!h]
\centering
    \subfloat[]{\includegraphics[width=0.35\linewidth]{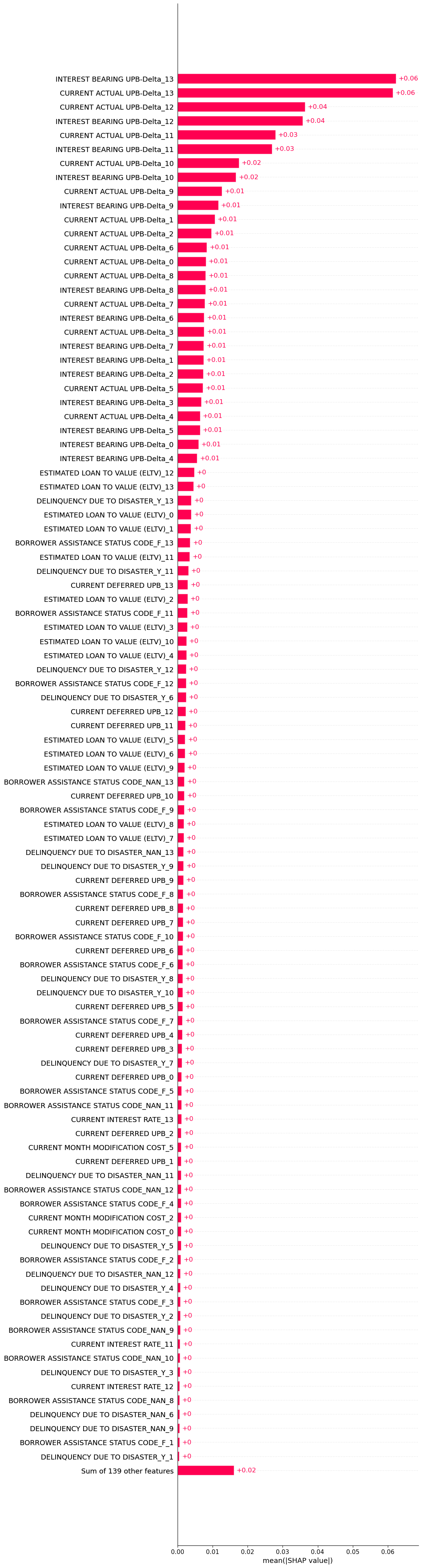} \label{fig:figure1f}}
    \subfloat[]{\includegraphics[width=0.35\linewidth]{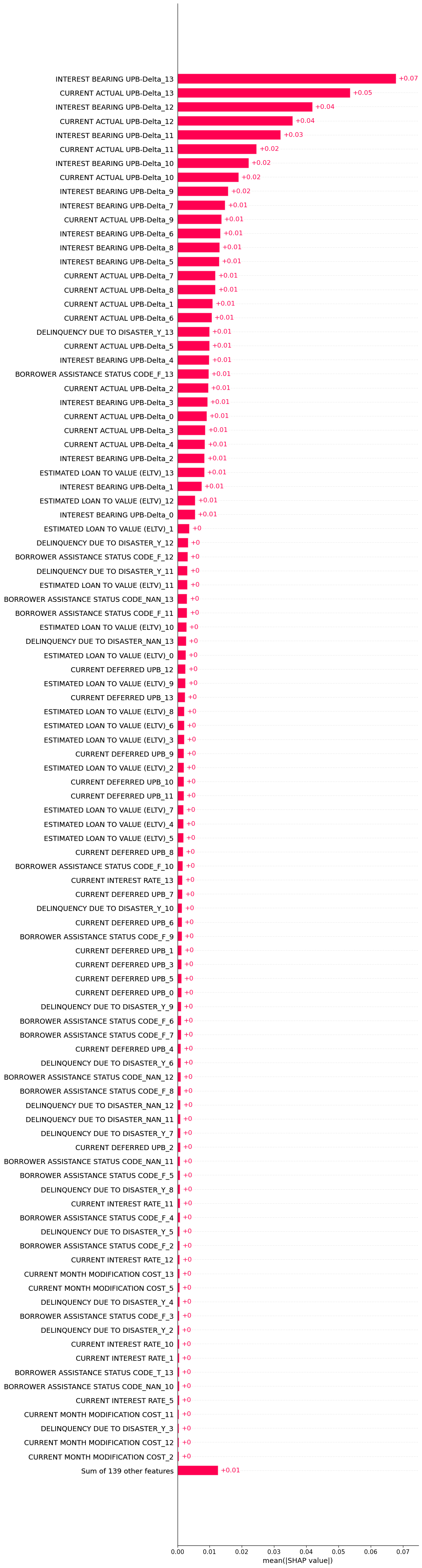} \label{fig:figure1a}}
    \subfloat[]{\includegraphics[width=0.35\linewidth]{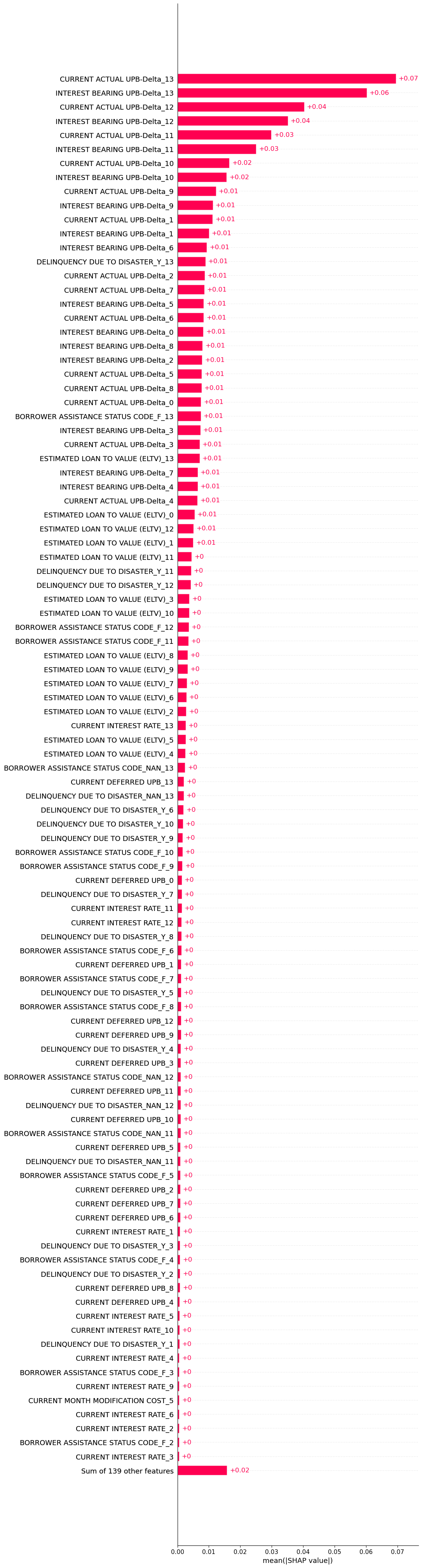} \label{fig:figure1d}}
    \caption{Barplot of (a) ResE-BiLSTM, (b) BiLSTM, and (c) LSTM. The vertical axis shows the importance rankings of the top 100 features in each month, with higher rankings indicating greater importance.}
\end{figure}

\begin{figure}[!h]
    \ContinuedFloat
    \centering
    \subfloat[]{\includegraphics[width=0.35\linewidth]{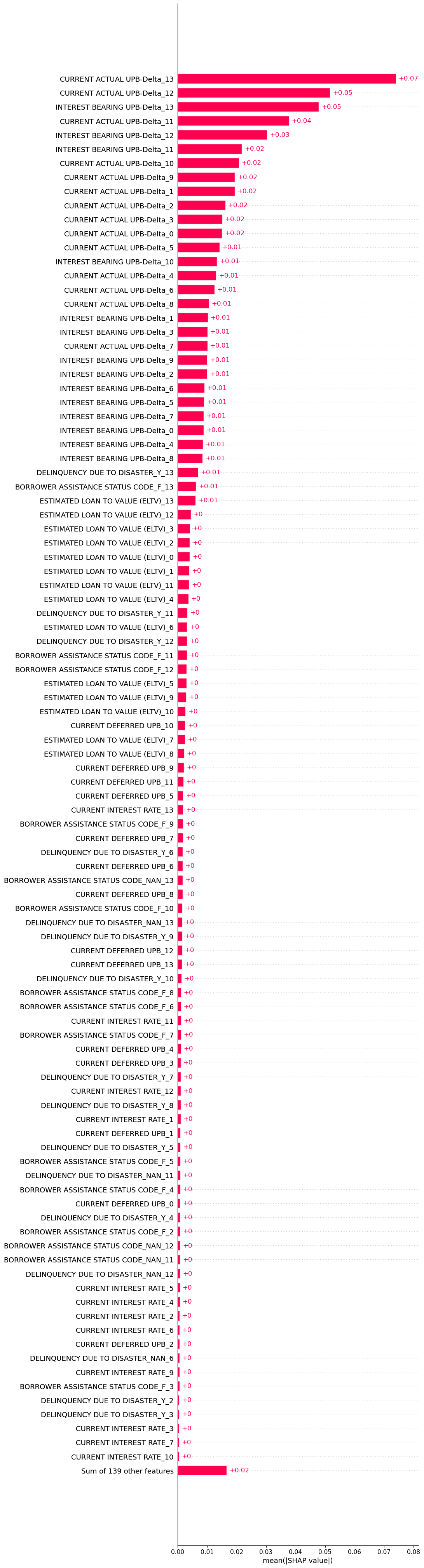} \label{fig:figure1c}}
    \subfloat[]{\includegraphics[width=0.35\linewidth]{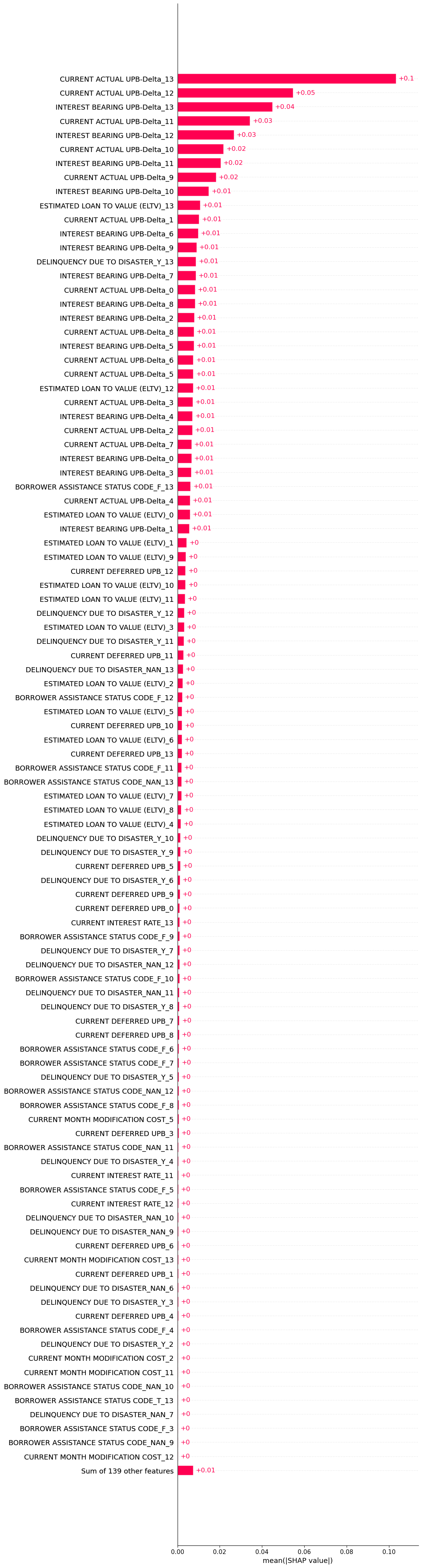} \label{fig:figure1e}}
    \subfloat[]{\includegraphics[width=0.35\linewidth]{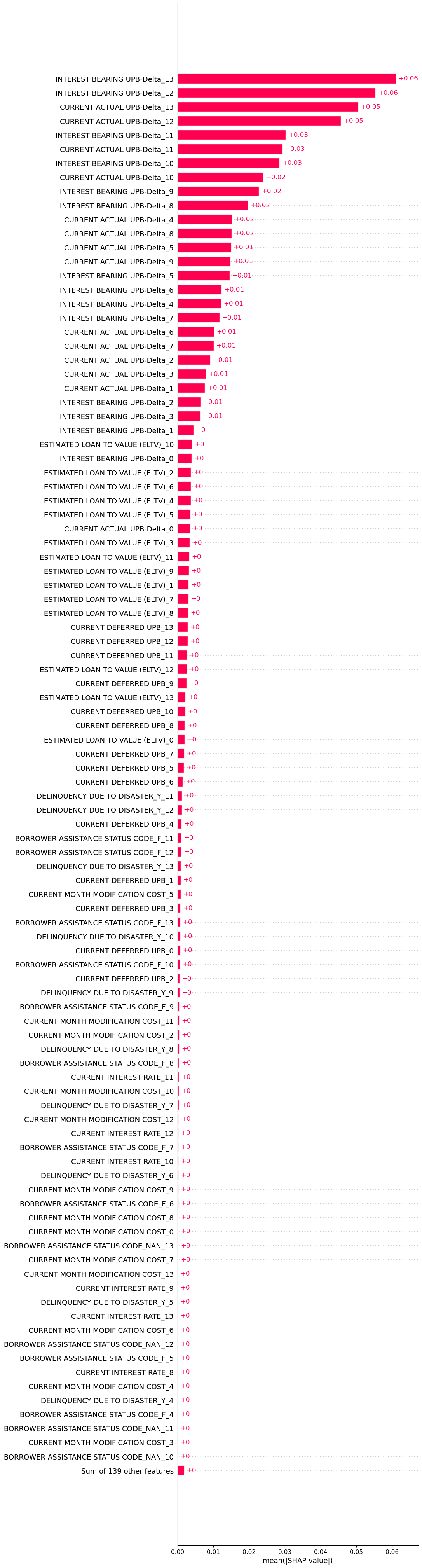} \label{fig:figure1b}}
    \caption{Barplot of (d) GRU, (e) RNN and (f) CNN. The vertical axis shows the importance rankings of the top 100 features in each month, with higher rankings indicating greater importance.}
\end{figure}

\begin{figure}[!h]
    \centering
    \subfloat[]{\includegraphics[width=0.23\linewidth]{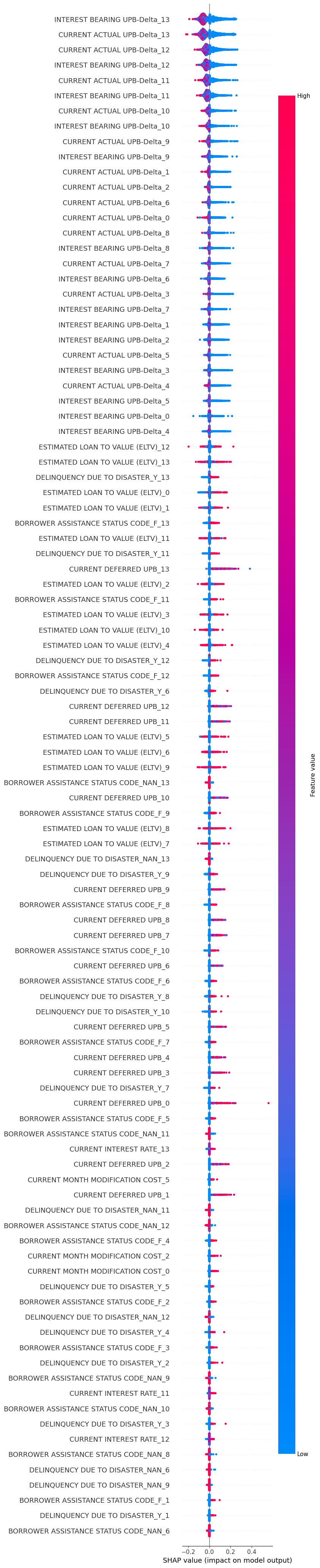} \label{fig:figure2f}}
    \subfloat[]{\includegraphics[width=0.23\linewidth]{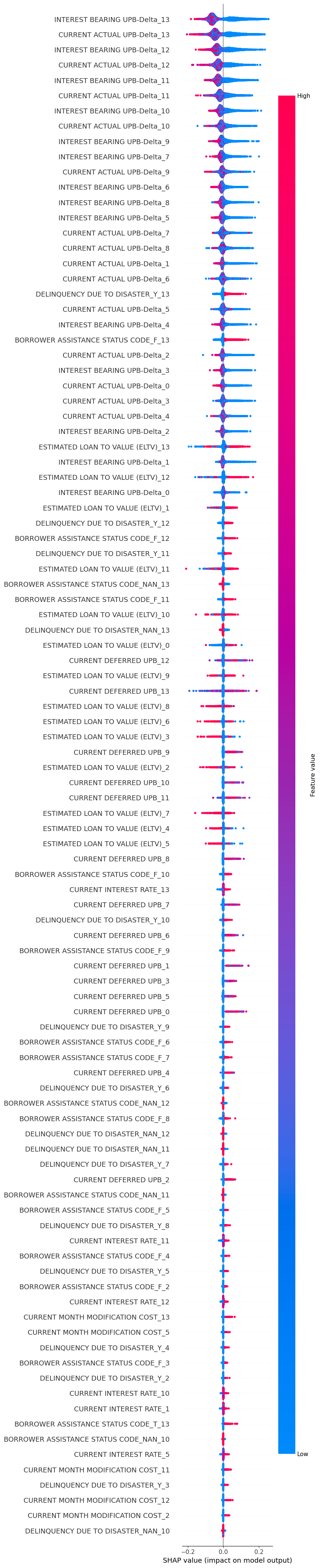} \label{fig:figure2a}}
    \subfloat[]{\includegraphics[width=0.23\linewidth]{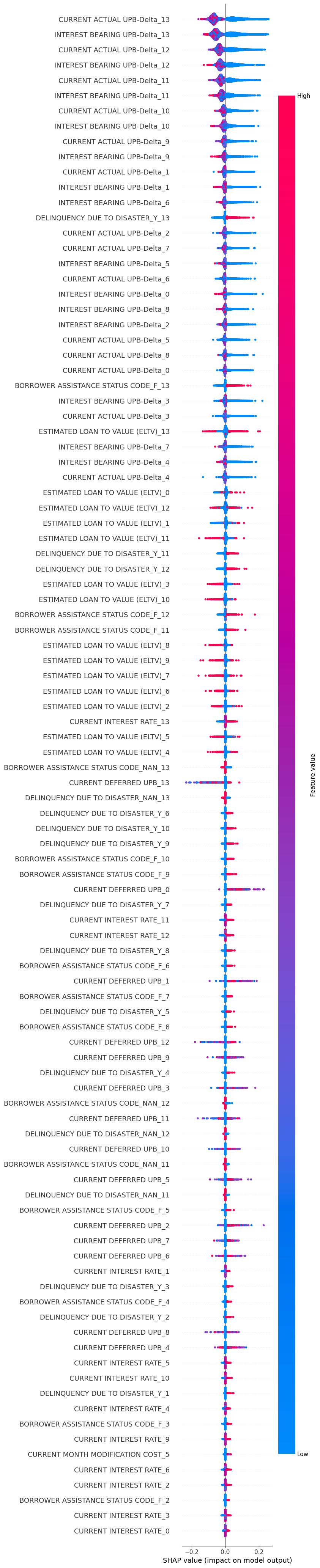} \label{fig:figure2d}}
    \caption{Summary plot of (a) ResE-BiLSTM, (b) BiLSTM, and (c) LSTM, showing dots for each sample. The vertical axis indicates sample density, while horizontal axis shows the SHAP value of the features. Rightward SHAP values suggest a stronger positive prediction impact, leftward, a stronger negative impact. Color coding reflects feature values, with red as high and blue as low.}
\end{figure}

\begin{figure}[!h]
    \ContinuedFloat
    \centering
    \subfloat[]{\includegraphics[width=0.23\linewidth]{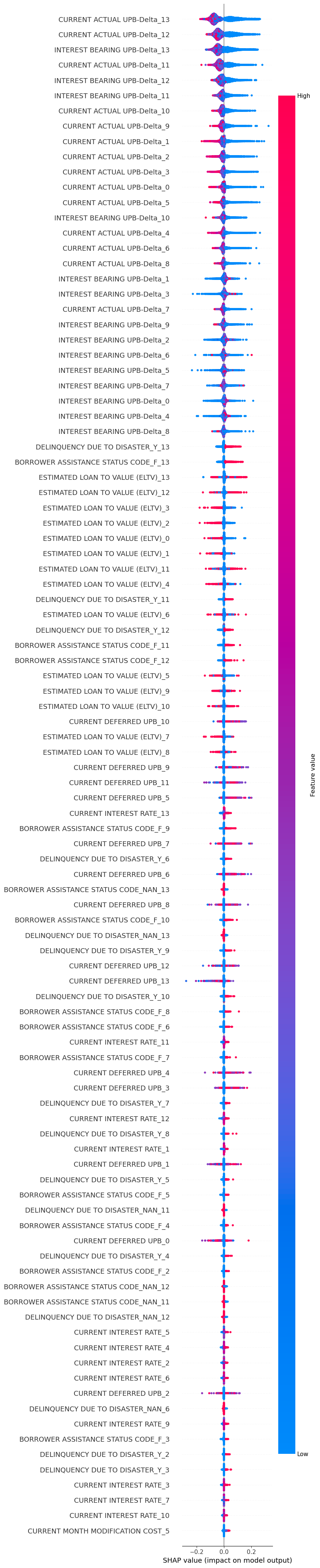} \label{fig:figure2c}}
    \subfloat[]{\includegraphics[width=0.23\linewidth]{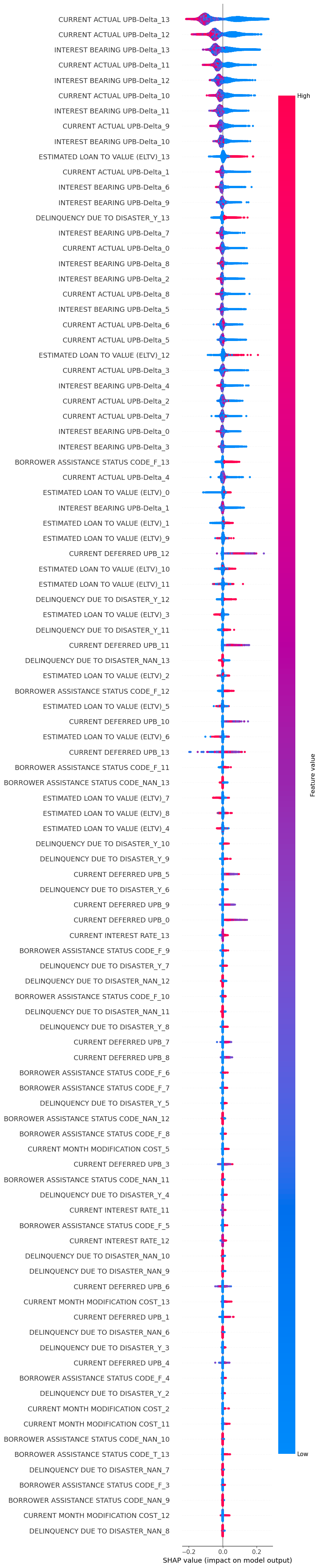} \label{fig:figure2e}}
    \subfloat[]{\includegraphics[width=0.23\linewidth]{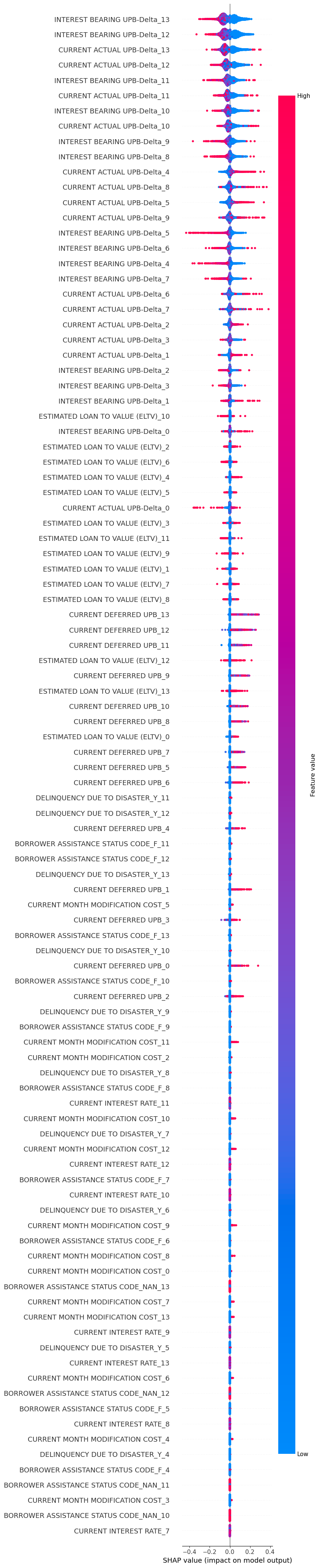} \label{fig:figure2b}}
    \caption{Summary plot of (d) GRU, (e) RNN and (f) CNN, showing dots for each sample. The vertical axis indicates sample density, while horizontal axis shows the SHAP value of the features. Rightward SHAP values suggest a stronger positive prediction impact, leftward, a stronger negative impact. Color coding reflects feature values, with red as high and blue as low.}
\end{figure}




\end{appendices}



\end{document}